\documentclass{article}

\usepackage{microtype}
\usepackage{graphicx}
\usepackage{booktabs} 
\usepackage{enumitem}

\usepackage[backref=page]{hyperref}

\usepackage{amsmath,amsfonts,bm}

\def\eqref#1{equation~\ref{#1}}

\def\1{\bm{1}}

\DeclareMathAlphabet{\mathsfit}{\encodingdefault}{\sfdefault}{m}{sl}
\SetMathAlphabet{\mathsfit}{bold}{\encodingdefault}{\sfdefault}{bx}{n}

\newcommand{\softmax}{\mathcal{S}}
\newcommand{\sigmoid}{\sigma}

\newcommand{\grad}{\nabla}

\newcommand{\loss}{\ell}
\newcommand{\barloss}{\bar{\ell}}

\newcommand{\vparam}{\boldsymbol{\theta}}
\newcommand{\param}{\theta}

\newcommand{\dkls}[3]{\mathbb{D}_{\text{KL}}^{#1}[#2 \, \|\, #3]}

\newcommand\cut[1]{}

\newcommand{\elbofinal}{\mathcal{L}}

\newcommand{\squishlist}{
	\begin{list}{$\bullet$}
		{ \setlength{\itemsep}{0pt}      \setlength{\parsep}{3pt}
			\setlength{\topsep}{3pt}       \setlength{\partopsep}{0pt}
			\setlength{\leftmargin}{1.5em} \setlength{\labelwidth}{1em}
			\setlength{\labelsep}{0.5em} } }
	
	\newcommand{\squishlisttwo}{
		\begin{list}{$\bullet$}
			{ \setlength{\itemsep}{0pt}    \setlength{\parsep}{0pt}
				\setlength{\topsep}{0pt}     \setlength{\partopsep}{0pt}
				\setlength{\leftmargin}{2em} \setlength{\labelwidth}{1.5em}
				\setlength{\labelsep}{0.5em} } }
		
		\newcommand{\squishend}{
	\end{list}  }
	
	{}
	\newtheorem{thm}{Theorem}{}
	
	{}
	{}

	\newenvironment{myproof}{{\bf Proof:}}{}

	\newcommand{\half}{\mbox{$\frac{1}{2}$}}

	\newcommand{\real}{\mbox{$\mathbb{R}$}}

	\newcommand{\rnd}[1]{\left(#1\right)}
	\newcommand{\sqr}[1]{\left[#1\right]}

	\newcommand{\myexpect}{\mathbb{E}}

	\newcommand{\gauss}{\mbox{${\cal N}$}}

\newcommand{\myvec}[1]{\mbox{$\mathbf{#1}$}}
\newcommand{\myvecsym}[1]{\mbox{$\boldsymbol{#1}$}}

\newcommand{\vepsilon}{\boldsymbol{\epsilon}}

\newcommand{\vphi}{\mbox{$\myvecsym{\phi}$}}

\newcommand{\vsigma}{\mbox{$\myvecsym{\sigma}$}}
\newcommand{\vSigma}{\mbox{$\myvecsym{\Sigma}$}}

\newcommand{\ve}{\mbox{$\myvec{e}$}}

\newcommand{\vf}{\mbox{$\myvec{f}$}}
\newcommand{\vg}{\mbox{$\myvec{g}$}}
\newcommand{\vh}{\mbox{$\myvec{h}$}}

\newcommand{\vm}{\mbox{$\myvec{m}$}}

\newcommand{\vu}{\mbox{$\myvec{u}$}}

\newcommand{\vy}{\mbox{$\myvec{y}$}}

\newcommand{\vI}{\mbox{$\myvec{I}$}}

\usepackage{subcaption}
\usepackage{url}

\usepackage[accepted]{icml2025}

\usepackage{amsmath}
\usepackage{amssymb}
\usepackage{mathtools}

\usepackage[capitalize,noabbrev]{cleveref}
\crefname{section}{Sec.}{Sections}
\crefname{appendix}{App.}{Appendices}
\crefname{algorithm}{Alg.}{Algorithms}
\crefname{equation}{Eq.}{Eqs.}
\crefname{figure}{Fig.}{Figures}
\creflabelformat{equation}{#2\textup{#1}#3} 

\let\classAND\AND
\let\AND\relax
\usepackage{algorithmic}
\usepackage{setspace}

\let\AND\classAND

\icmltitlerunning{Variational Learning Induces Adaptive Label Smoothing}

\begin{document}

\twocolumn[
\icmltitle{Variational Learning Induces Adaptive Label Smoothing}



\icmlsetsymbol{equal}{*}

\begin{icmlauthorlist}
\icmlauthor{Sin-Han Yang}{yyy}
\icmlauthor{Zhedong Liu}{yyy}
\icmlauthor{Gian Maria Marconi}{yyy,wov}
\icmlauthor{Mohammad Emtiyaz Khan}{yyy}
\end{icmlauthorlist}

\icmlaffiliation{yyy}{RIKEN Center for Advanced Intelligence Project, Tokyo, Japan}
\icmlaffiliation{wov}{Woven by Toyota}

\icmlcorrespondingauthor{Sin-Han Yang}{sin-han.yang@riken.jp}
\icmlcorrespondingauthor{Mohammad Emtiyaz Khan}{emtiyaz.khan@riken.jp}

\icmlkeywords{Machine Learning, ICML}

\vskip 0.3in
]



\printAffiliationsAndNotice{}  

\begin{abstract}
    We show that variational learning naturally induces an adaptive label smoothing where label noise is specialized for each example. Such label-smoothing is useful to handle examples with labeling errors and distribution shifts, but designing a good adaptivity strategy is not always easy. We propose to skip this step and simply use the natural adaptivity induced during the optimization of a variational objective. We show empirical results where a variational
    algorithm called IVON outperforms traditional label smoothing and yields adaptivity strategies similar to those of an existing approach.~By connecting Bayesian methods to label smoothing, our work provides a new way to handle overconfident predictions.
 \end{abstract}
 
 \section{Introduction}
 
 Adaptive strategies to Label Smoothing (LS) \citep{Szegedy_rethinking} aim to adapt the label noise according to the type of data example. Such adaptation can be more effective in practice than its traditional counterpart where the label noise is the same for all examples. Adaptation is useful to handle examples that may have labeling errors, distribution shift, or calibration issues. For such cases, the effectiveness of adaptation has been extensively
 studied, for example, see \citet{ghoshal2021learning,lee2022adaptive} for 
 generalization improvements, \citet{zhang2021delving,ko2023gift} for mislabelled examples, \citet{park2023acls} for miscalibration, and \citet{xu2024adaptive} for out-of-distribution detection. Adaptivity is useful for label smoothing to handle all such cases.

One major problem with adaptive label smoothing is that it is not easy to design a good adaptivity strategy.
For example, a simple approach is to adapt the label noise by using model's predictions but there are many ways to do this, for examples, \citet{park2023acls} set the noise based on the logits, 
 \citet{zhang2021delving,ko2023gift} use the predictive probabilities (obtained with softmax), while \citet{lee2022adaptive} use their entropy.
All of these are reasonable ideas but the choice of a good strategy for a given problem is not always straightforward. A strategy that reduces miscalibration may not be most effective for handling outliers or mislabeling. Focusing on one issue at a time has given rise to a lot of ad-hoc and heuristic strategies, and, despite their usefulness, designing an adaptive strategy for a task in hand remains tricky.
Our goal here is to simplify the process by presenting and analyzing algorithms that naturally induce adaptivity. 
 
\begin{figure}[!t]
	\centering
      \includegraphics[width=\linewidth]{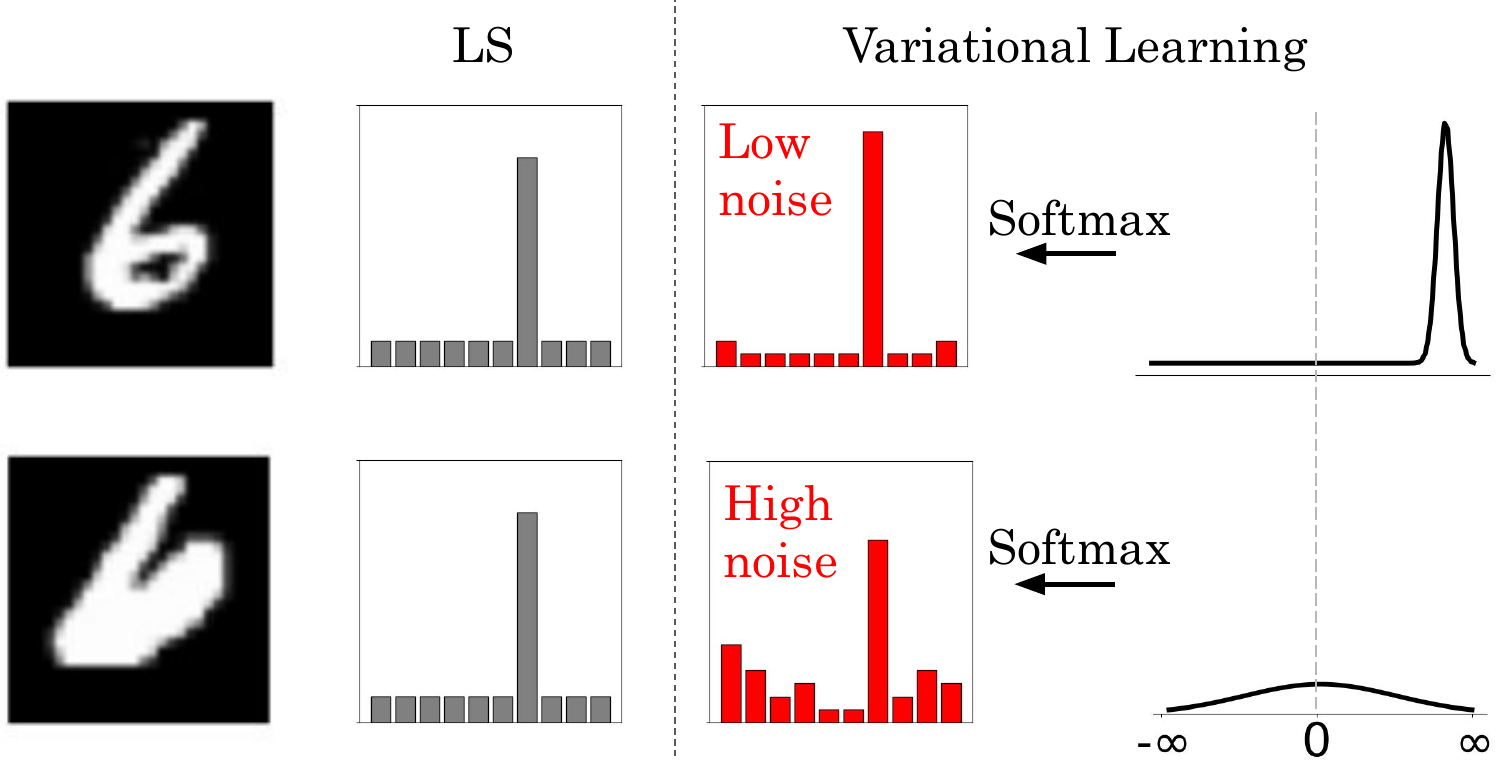}
   \caption{Given a regular $6$ digit (top) and an atypical one (bottom), Label Smoothing (LS) assigns the same label noise to both (gray bars) while variational learning assigns higher noise to the atypical example (red bars). Adaptivity naturally arises due to the posterior.}
   \label{fig:illustration}
\end{figure}

We show that variational learning naturally induces an adaptive label smoothing. The smoothing arises due to the use of the expectation of the loss in the variational objective, taken with respect to the posterior distribution. The expectation gives rise to a label noise (among other types of noises) which is customized for each example through its features.
 Our key contribution is to derive the exact form of the label noise (\cref{eq:labelnoiseGD}) for many problems and study their behavior.
 We show extensive empirical results analyzing the label noise induced by Improved Variational Online Newton (IVON) \citep{IVON}. We show the following: 
 \begin{enumerate}[itemsep=-0.4pt]
    \item Variational learning assigns higher noise to atypical or ambiguous examples (\cref{fig:illustration} and \cref{fig: mnist_noise_distribution}).
    \item IVON's adaptive label noise behaves similarly to the proposal of \citet{zhang2021delving}.
    \item IVON consistently outperforms Label Smoothing in presence of labeling errors,  giving up to 9\% accuracy boost for pair-flip noise (\cref{fig:cifar100_all}) and sometimes even around 50\% for data-dependent noise (\cref{fig:cifar10_inc_acc}). 
 \end{enumerate}
Our work connects label smoothing literature to Bayesian methods, thereby providing a new way to handle overconfident predictions in deep learning.
 
\section{Label Smoothing and Adaptivity Strategies} \label{sec: label smoothing}
 
 Label Smoothing (LS) is a simple technique where the true label vector $\vy_i$ (length $K$) are replaced by a smoothed version. In its simplest form, a convex combination is used where the smoothed labels are defined as
 \begin{equation} \label{equ: label smoothing def}
   \vy_i'=(1-\alpha)\vy_i + \alpha \vu,
 \end{equation}
for some scalar $\alpha \in (0, 1)$ with $\vu$ as a vector of $1/K$ with $K$ being the number of classes. This simple technique is effective to penalize overconfident predictions because the noise $\alpha(\vu -\vy_i)$ reduces the importance of the label during training \citep{pereyra2017regularizing}. Multiple works have studied its effectiveness, for example, to improve calibration and representation \citep{muller2019does}, to favor flatter
 solutions \citep{damian2021label}, and improve robustness to mislabelled data \citep{lukasik2020does,liu2021understanding} due to its connections to loss correction \citep{patrini2017making}. Despite its simplicity, LS has clear practical advantages.
 
 
Adaptive label smoothing aims to inject noise according to the type of data example, for example, during learning, we may want to inject a noise to get the smoothed label
\begin{equation}
   \vy_{i|t} = \vy_i + \vepsilon_{i|t}.
\end{equation}
The noise $\vepsilon_{i|t}$ depend on the model parameter $\vparam_t$ at iteration $t$, and can be varied according the model's opinion regarding the relevance of the examples. Adaptive label smoothing uses additive noise to reweigh examples during training.
Many studies have shown the effectiveness of the adaptive noise, which ranges from improvements in generalization \citep{ghoshal2021learning,lee2022adaptive}, robustness to mislabeled data \citep{zhang2021delving,ko2023gift}, improving calibration \citep{park2023acls} and out-of-distribution (OOD) detection \citep{xu2024adaptive}. By adapting label noise, such methods aim to down-weight the problematic examples.


While adaptivity is desirable, it also requires additional effort to design a good strategy to adapt. Each specific issues may require a different type of noise, for instance, what works to reduce miscalibration, may not be most effective for handling OOD detection or mislabling. Focusing on one issue or strategy at a time has given rise to a lot of ad-hoc and heuristic strategies, and, despite their usefulness, clarity of good ways to design adaptive strategy is lacking.

The simplest approach is to adapt by using the model predictions based on the logits $\vf_i(\vparam_t)$, but there are many ways to use them. \citet{zhang2021delving} use the following update for each example $i$ in the epoch $t$
\begin{equation}
    \vu = \sum_i \softmax\sqr{ \vf_i(\vparam_t)},
\end{equation}
 where $\softmax[\vf]$ vector (length $K$) with $j$'th entry defined as
 \begin{equation}
    \softmax_j \sqr{ \vf} = \frac{ e^{f_j}}{ \sum_{k=1}^K e^{f_k} }.
    \label{eq:softmax}
 \end{equation}
The noise injected by this method is
\begin{equation} \label{equ: ols noise}
   \vepsilon_{i|t} = \alpha (\bar{\vu}-y_i),
\end{equation}
where $\bar{\vu}$ is the normalized $\vu$. A similar rule is used by \citet{ko2023gift}. Instead of directly using the logits, \citet{lee2022adaptive} use them to adjust $\alpha$. They do so by using the entropy of the model-output distribution, assigning a smaller smoothing to high entropy samples and larger smoothing to low entropy samples. Another approach by \citet{park2023acls} decrease the label noise linearly as the logit $\vf_i(\vparam_t)$ increase.
There are multiple ways to use predictions but the choice of a good strategy for a given problem is not always straightforward.

Intuitively, using model's predictions makes sense because predictions can tell us about the relevance of examples. Regions where model is inaccurate may also contain examples that need special attention but also those that are impossible to predict. Some works have explored this from the Bayesian viewpoint, although only using the posterior over the labels. For example, \citet{li2020regularization} motivate adaptive smoothing using Bayes error rate, which
implies larger smoothing to example that lie near the decision boundary. Similarly, \citet{ghoshal2021learning} use a PAC-Bayes bound to motivate adaptivity. However, there are no approaches investigating the effectiveness of posterior over $\vparam$.

In this paper, we show that directly learning the posterior using a variational method natural yields an adaptive label noise. Adaptivity introduced in this fashion directly takes various causes of uncertainty, some of which are then handled through the label noise. The uncertainty in parameter have other desired effect that are often missed when only focusing on the label noise. In our context, this can simplify the design of adaptive label smoothing or may even allow us to
entirely skip the step. We will now discuss the adaptive label noise induced by variational learning.

\section{Variational Learning Induces Adaptive LS}

Variational learning aims to optimize for distribution over parameters $\vparam$ which is fundamentally different from traditional deep learning where we minimize empirical risk,
\begin{equation}
   \barloss(\vparam) = \sum_{i=1}^N \loss_i(\vparam) + \mathcal{R}_0(\vparam),
   \label{eq:dl}
\end{equation}
with loss $\loss_i(\vparam)$ for the $i$'th example in the training dataset.
The regularizer $\mathcal{R}_0(\param)$ is often implicitly defined through various training choices, such as, weight-decay, initialization, and architecture design. In contrast, variational learning aims to find a distribution $q(\vparam) \in\mathcal{Q}$ which minimizes
\begin{equation}
   \elbofinal(q) = \sum_{i=1}^N \myexpect_q \sqr{ \loss_i(\vparam)} + \dkls{}{q(\vparam)}{p(\vparam)}.
   \label{eq:vl}
\end{equation}


The second term is the Kullback-Leibler (KL) Divergence where the $p(\vparam) \propto \exp(-\mathcal{R}_0(\vparam))$ can be defined implicitly similarly to deep learning. Throughout, we will set $q(\vparam)$ to take Gaussian forms and show that, despite their differences, variational learning can be implicitly seen as minimizing a noisy version of \cref{eq:dl}. Existing works have studied the weight-noise \citep{zhang2018noisy,khan2018fastadam} but our goal here is to specifically
study its effect on label noise. 

\subsection{A Simple Example: Logistic Regression}
We start with logistic regression where we can write a closed-form expression for the adaptive label noise. The result extends to all loss functions using generalized linear model. We will consider all such extensions (including neural networks) afterwards. For now, we consider a loss function for binary labels $y_i \in \{0,1\}$ with model output $f_i(\vparam)$,
\begin{equation}
   \ell_i(\vparam) = -y_i f_i(\vparam) + \log \rnd{1+e^{f_i(\vparam)}}.
   \label{eq:bce}
\end{equation}
In logistic regression, we have $f_i(\vparam) = \vphi_i^\top\vparam$ where $\vphi_i \in\real^P$ is the feature vector. For simplicity, let us assume $\mathcal{R}_0(\vparam) = \half \|\vparam\|^2$ to be a quadratic regularizer. 
For such a model, we can solve \cref{eq:dl} with gradient descent (GD),
\begin{equation}
   \vparam_{t+1} = (1-\rho_t)\vparam_t - \rho_t \sum_{i=1}^N \vphi_i \sqr{ \sigmoid( f_i( \vparam_t) ) - y_i }   
   \label{eq:gd}
\end{equation}
The result is obtained by simply taking the derivative of \cref{eq:bce} which gives rise to $\sigmoid(f) = 1/(1+e^{-f})$, a binary version of the softmax function from \cref{eq:softmax}.
We will now show that, by choosing the family $\mathcal{Q}$ appropriately, variational learning can be seen as GD with label noise.

We choose the distribution $q_t(\vparam)$ at iteration $t$ to take a Gaussian form with mean $\vparam_t$ and covariance set to the identity,
\[
   q_t(\vparam) = \gauss(\vparam | \vparam_t, \vI),
\]
and perform GD to minimize the variational objective in \cref{eq:vl}, now denoted as $\elbofinal(\vparam_t)$, with respect to $\vparam_t$. 
Below is a formal statement of the result.
\begin{thm}
   A gradient update $\vparam_{t+t} = \vparam_t - \rho_t \nabla_{\vparam_t} \mathcal{L}(\vparam_t)$ is equivalent to the gradient update in \cref{eq:gd} where the label $y_i$ are replaced by $y_i +\epsilon_{i|t}$ with noise defined as 
   \begin{equation}
      \epsilon_{i|t} =  \sigmoid(f_i(\vparam_t)) - \myexpect_{q_t} [\sigmoid(f_i(\vparam))].
      \label{eq:labelnoiseGD}
   \end{equation}
\end{thm}

\begin{myproof}
The gradient of the expected loss in \cref{eq:vl} can be simplified to take a form very similar to the one in \cref{eq:gd},
\begin{equation}
   \begin{split}
      \nabla_{\vparam_t} \myexpect_{q_t} [ \loss_i(\vparam)] &= \nabla_{\vparam_t} \myexpect_{\text{\gauss}(\text{\ve}|0, \text{\vI})} [ \loss_i(\vparam_t + \ve)] \\
      &= \myexpect_{\text{\gauss}(\text{\ve}|0, \text{\vI})} \sqr{ \nabla_{\vparam_t} \loss_i(\vparam_t + \ve) } \\
      &= \vphi_i  \sqr{ {\myexpect_{q_t}} [\sigmoid( f_i( \vparam) )] - y_i } 
   \end{split}
   \label{eq:expected_grad}
\end{equation}
The gradient of KL is also simplifies to
\[
   \dkls{}{q(\vparam)}{p(\vparam)} = \myexpect_q\sqr{ \log \frac{q(\vparam)}{p(\vparam)} } = \half \|\vparam\|^2 + \text{const.}
\]
   Using these, we can write the GD to minimize \cref{eq:vl} as
\[
   \vparam_{t+1} = (1-\rho_t)\vparam_t - \rho_t \sum_{i=1}^N \vphi_i \sqr{ {\color{red} \myexpect_{q_t}} [\sigmoid( f_i( \vparam) )] - y_i } ,
\]
   which has a similar form as \cref{eq:gd} but with one difference: $\sigmoid( f_i( \vparam) )$ are replaced by their expectation over $q$ (highlighted in red).
   By adding and subtracting $\sigmoid( f_i( \vparam_t) )$, we can rewrite the update as \cref{eq:gd} which has the label noise defined in \cref{eq:labelnoiseGD}. $\hfill\blacksquare$
\end{myproof}

\begin{figure}[!t]
	\centering
   \includegraphics[width=0.85\linewidth]{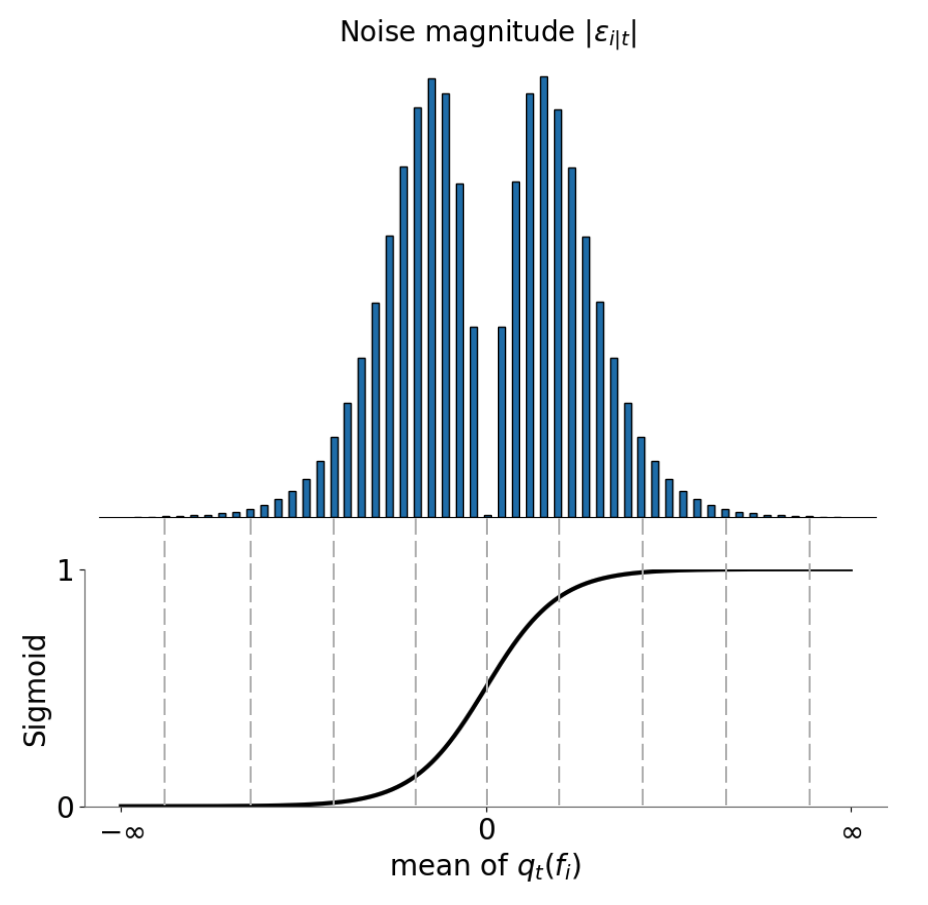}
   \caption{We plot label noise magnitude $\epsilon_{i|t}$ from \cref{eq:labelnoiseGD_sim} by varying the mean $f_{i|t}$ of $q_t(f_i)$ while fixing its variance to 1. The noise is large around 0 (but not at 0) with large peaks on both sides.}
   \label{fig: guassian after sigmoid}
\end{figure}

The result shows that the GD steps to optimize \cref{eq:vl} is equivalent to those to optimize \cref{eq:dl} but with a noisy label. The noise is adaptive and depends on where the Gaussian distribution is located. To show this, we derive the distribution over $f_i = \vphi_i^\top\vparam$, which takes a Gaussian form: 
\begin{equation}
   q_t(f_i) = \gauss(f_i|f_{i|t}, \vphi_i^\top\vphi_i),
   \label{eq:qf_logreg}
\end{equation}
where we denote $f_{i|t} = \vphi_i^\top\vparam_t$. The label noise then is simply the difference between the sigmoid $\sigmoid(f_{i|t})$ of the mean $f_{i|t}$ and mean of $\sigmoid(f_i)$ with respect to $q_t(f_i)$, that is
\begin{equation}\label{eq:labelnoiseGD_sim}
   \epsilon_{i|t} = \sigmoid(f_{i|t}) - \myexpect_{q_t}[\sigmoid(f_i)],
\end{equation}
where the form is similar to the noise of \citet{zhang2021delving} in \cref{equ: ols noise}. \cref{fig: guassian after sigmoid} plots the magnitude of this quantity as a function of the mean $f_{i|t}$ but fixing the variance $\vphi_i^\top\vphi_i = 1$. We see the noise to be large whenever $f_{i|t}$ around 0, with the maximum in areas slightly away from it. The $\sigmoid(f)$ is flat far away from 0 and uncertainty in $q_t$ is amplified around 0, which makes the difference large around 0 (but not at 0).

The other factor that affects the noise is the feature $\vphi_i$. Inputs with larger features induce larger variance. When the features are normalized, this is unlikely to have an effect, but this is important for the neural networks case where features
are learned. 

The two factors explain why we would expect high label noise for atypical or ambiguous examples. This is because the predictive distribution $q(f_{i|t})$ is close to 0 and may also have a higher variance. An alternate way to understand the impact of the two factors is to use a Taylor's approximation at a sample $e \sim \gauss(0,1)$,
\begin{equation}
   \epsilon_{i|t} \approx \sigmoid(f_{i|t}) - \sigmoid(f_{i|t} + \|\vphi_i\|_2 e)  \approx  \sigmoid'(f_{i|t}) (\vphi_i^\top \vphi_i)^{1/2} e .
   \label{eq:taylor_noise}
\end{equation}
We again see the two factors: one is $\sigmoid'(f)$ (which peaks around 0) and the other is the feature norm.~Note that this approximation does not get better for larger number of samples, but it roughly captures the behavior away from 0.

\subsection{Generalized Linear Model (GLM) with GD}

The result generalizes to any loss function derived using exponential-family distribution, for instance, the following generalization of \cref{eq:bce}
\begin{equation} \label{eq:glm_loss}
   \ell_i(\vparam) = -\vy_i^\top \vf_i(\vparam) + A(\vf_i(\vparam) ),
\end{equation}
where $A(\vf)$ is a convex function called the log-partition function.
The regularizer can also be a general convex function. For such models, we can derive the label noise following almost the same procedure as in the previous section. Due to its similarity, we omit the derivation and only give the final form of the noise,
\begin{equation}
   \vepsilon_{i|t} = A'(\vf_i(\vparam_t)) - \myexpect_{q_t} [A'(\vf_i(\vparam))].
   \label{eq:labelnoiseGLMGD}
\end{equation}
Essentially, we replace the $\sigmoid(f)$ by the derivative $A'(f)$. For logistic regression, $A(f) = \log (1+ e^f)$, derivative of which is $\sigmoid(f)$ and we recover the result in \cref{eq:labelnoiseGD}. We can extend this result to multiclass classification by considering $A(\vf) = \log \sum_{k=1}^K e^{f_k}$, derivative of which is the softmax function defined in \cref{eq:softmax}. Similarly to the binary case, we expect uncertainty in $q_t$ to be amplified near the
boundary. The label noise is therefore low for examples where softmax yields probabilities close to 0 or 1.

\subsection{Generalized Linear Model with Newton's Method} \label{sec: newton on GLM}

We now go beyond GD to Newton's method and show that a specific variational-learning algorithm can be seen as a noisy-label version of Newton's method. This is a useful step before we move to neural networks training. Here, we find that the form of the noise has exactly same form as \cref{eq:labelnoiseGLMGD} but the distribution $q_t$ has a flexible covariance which improves the adaptivity of the label noise. 

We consider the following Newton's update,
\begin{equation}
   \vparam_{t+1} = \vparam_t - \sqr{ \nabla^2 \barloss(\vparam_t) }^{-1} \nabla \barloss(\vparam_t)
   \label{eq:newton}
\end{equation}
which is commonly used for generalized linear models. As shown by \citet{khan2023bayesian}, the update can be seen as a special case of a Variational Online Newton (VON) algorithm \citep{khan2018fastadam} to learn a full Gaussian with covariance $\vSigma_t$,
\[
   q_t(\vparam) = \gauss(\vparam | \vparam_t, \vSigma_t) 
\]
The VON updates are given as follows,
\begin{equation}
   \begin{split}
      \vparam_{t+1} &= \vparam_t - \rho_t \vSigma_{t+1} \myexpect_{q_t} [ \nabla \barloss(\vparam)] \\
      \vSigma_{t+1}^{-1} &= (1-\rho_t) \vSigma_{t}^{-1} + \rho_t \myexpect_{q_t} [ \nabla^2 \barloss(\vparam)] .
   \end{split}
   \label{eq:von}
\end{equation}
Setting $\rho_t = 1$ yields a Newton-like update where gradients $\nabla\barloss$ and Hessian $\nabla^2\barloss$ are replaced by their terms where expectations are taken, namely, $\myexpect_{q_t}[\nabla\barloss]$ and $\myexpect_{q_t}[\nabla^2 \barloss]$. Similarly to the previous cases, the label noise in VON arises due to the expectation of the gradient, while expectation of the Hessian gives rise to other types of noise.

As shown in in \cref{app:glm_newton}, the VON updates in \cref{eq:von} are equivalent to Newton's update in \cref{eq:newton} where labels are replaced by the noisy ones with noise shown in \cref{eq:labelnoiseGLMGD}. The proof technique relies on comparing the form of the surrogates for the two algorithms. Even though the noise has the same form, there is an important difference here. Essentially, the Gaussian $q_t$ now is more flexible because its covariance $\vSigma_t$ is not fixed but learned using the Hessian. As a result the distribution over $f_i$ now has adaptive variances, 
\begin{equation}
   q_t(f_i) = \gauss(f_i|f_{i|t}, \vphi_i^\top\vSigma_t\vphi_i).
   \label{eq:qf_newton}
\end{equation}
Therefore, now both the location and spread of the Gaussians are changed for each example, and they both contribute to the adaptivity. The result shows that second-order methods yield more adaptive label noise than first order methods, and are expected to perform better in practice. We will later present experiments that support this finding. 

\subsection{Neural Network training with IVON} \label{sec: IVON on NN}

\begin{figure*}[h!]
	\centering
	\begin{subfigure}{0.45\linewidth}
		\centering
		\includegraphics[width=\linewidth]{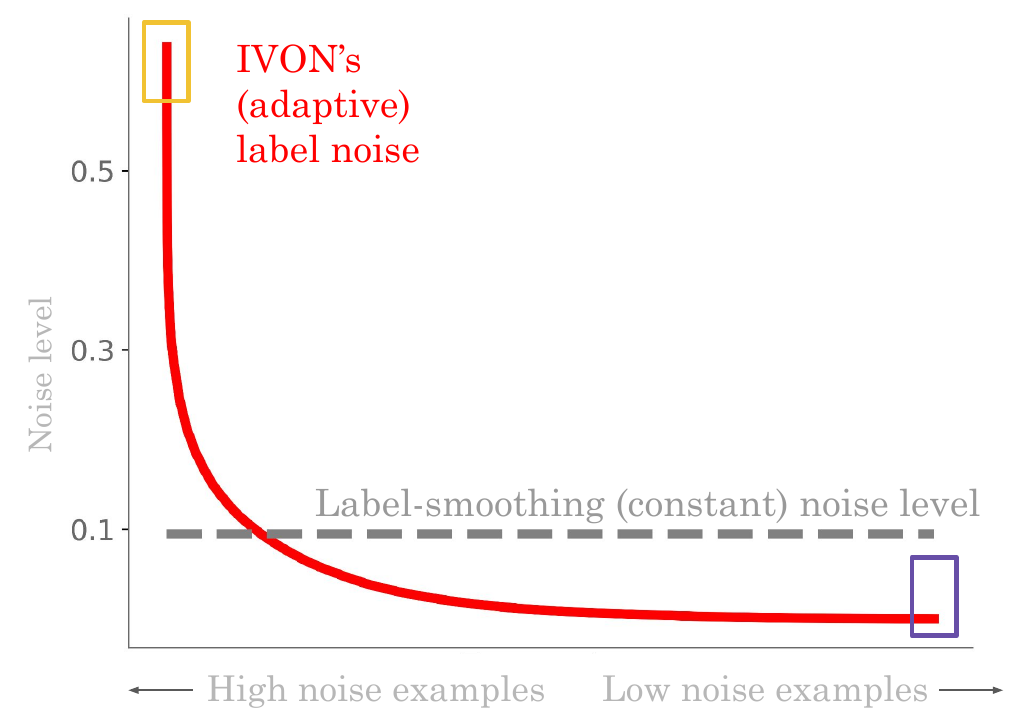}
	\end{subfigure} 
	\begin{subfigure}{0.44\linewidth}
		\centering
		\includegraphics[width=\linewidth]{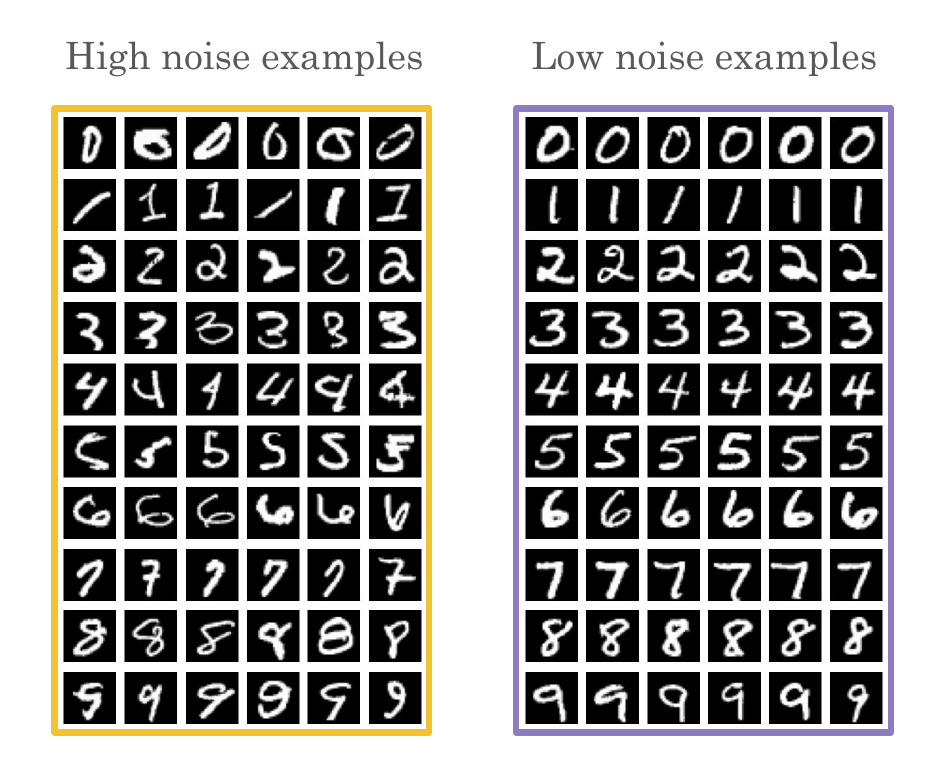}
	\end{subfigure}
	\caption{Label noise assigned by IVON and LS in MNIST dataset. Examples are ordered according to IVON's noise, and highest and lowest noise examples are visualized. We see that high noise is assigned to atypical examples while low noise is assigned to regular ones.}
	\label{fig: mnist_noise_distribution}
\end{figure*}

We will now show that the label noise expression have similar form for the neural network case, but to derive them we need to use Taylor's approximation. Essentially, the form of the expression then is similar to \cref{eq:taylor_noise} there the adaptive nature should roughly stay the same. We validate these findings later through numerical experiments.

We will illustrate the derivation for the binary case which can then be extended to other case as we did in previous section. Taylor's approximations is required because the gradient of $\loss_i$, shown below,
\[
   \nabla \loss_i(\vparam) = \nabla f_i(\vparam_t) \sqr{ \sigmoid( f_i( \vparam_t) ) - y_i },
\]
replaces the $\vphi_i$ term in \cref{eq:gd} by $\nabla f_i(\vparam_t)$. As a result, we cannot simply move the expectation over $q_t$ to derive the label noise as we did in \cref{eq:expected_grad}. However, we can simplify these by using Taylor's approximation.

We show this by using a single-sample $\vparam_t^{(1)} \sim q_t$ Monte-Carlo approximation (multiple samples can also be used),
\[
   \myexpect_{q_t} \sqr{ \nabla \loss_i(\vparam)} 
   \approx \nabla f_i(\vparam_t^{(1)}) \sqr{ \sigmoid( f_i( \vparam_t^{(1)}) ) - y_i } .
\]
Then, we do the following two approximations where we use Taylor's expansion but ignore the second-order terms, 
\begin{align*}
   \sigmoid(f_i(\vparam_t^{(1)})) &\approx \sigmoid(f_i(\vparam_t)) + \sigmoid'(f_i(\vparam_t)) \nabla f_i(\vparam_t) (\vparam_t^{(1)} - \vparam_t) \\
   \nabla f_i(\vparam_t^{(1)}) &\approx \nabla f_i(\vparam_t) 
\end{align*}
With these approximations, we can write,
\[
   \myexpect_{q_t} \sqr{ \nabla \loss_i(\vparam)} \approx \nabla f_i(\vparam_t) \sqr{ \sigmoid( f_i( \vparam_t) ) - (y_i + \epsilon_{i|t})},
\]
where the noise takes a very similar form to \cref{eq:taylor_noise},
\begin{equation}
   \epsilon_{i|t} \approx \sigmoid'(f_i(\vparam_t)) \nabla f_i(\vparam_t) \vSigma_t^{1/2} \ve
   \label{eq:taylor_noise_nn}
\end{equation}
where $\ve$ is a sample from a standard normal distribution. The derivation generalizes to all GLM losses by replacing $\sigmoid(\cdot)$ by $A'(\cdot)$. It also extends to variational GD and VON.

In practice, neural networks are trained with Adam-style algorithm. In our experiments, we will use an Adam-like version of VON, called IVON, which is recently proposed by \citet{IVON}. The key different to VON is that it estimates a diagonal covariance by using an Adam-like preconditioning update; a pseudo-code is added in \cref{alg:ivon}. The diagonal covariance is estimated through the scale vector. We will use the label noise expression given in \cref{eq:taylor_noise_nn} where $\vSigma_t$ is replaced by the
diagonal covariance estimated by IVON. Note that variational learning for neural neworks with IVON introduces many other noise other than label noise, for instance, the noise is introduced in the features $\nabla f(\vparam)$, as shown above. We will analyze only the label noise but the performance is affected by other noises too.

In our experiments, we also compare to Sharpness-Aware Minimization (SAM) \citep{SAM_org} which has a variational interpretation \citep{thomassampaper} and has been shown to perform well with mislabelled data. Using our techniques, it is possible to derive the label noise of SAM but the expression would be similar to the one derived here.
The difficulty with SAM is that we need to tune the `size' of adversarial perturbation, often denoted by a scalar $\rho$, while IVON can automatically estimate it using the posterior variance. In our experiments, we show that IVON performs comparably to SAM with a highly tuned $\rho$, and it does not need to set any such hyperparameters.

 \section{Experiments}

\begin{figure*}[t!]
	\centering
	\begin{subfigure}{0.33\linewidth}
		\centering
		\includegraphics[width=\linewidth]{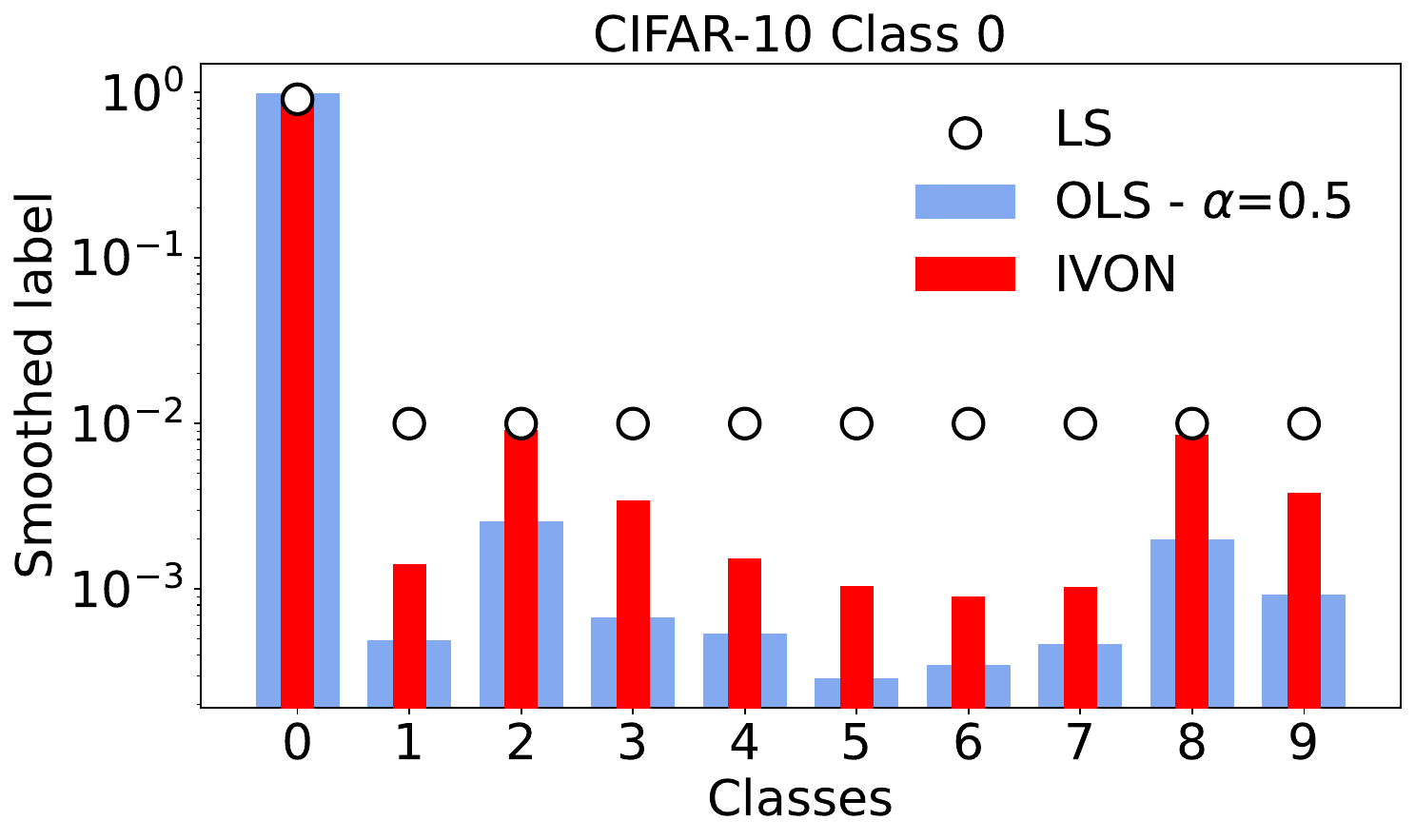}
	\end{subfigure}
	\begin{subfigure}{0.33\linewidth}
		\centering
		\includegraphics[width=\linewidth]{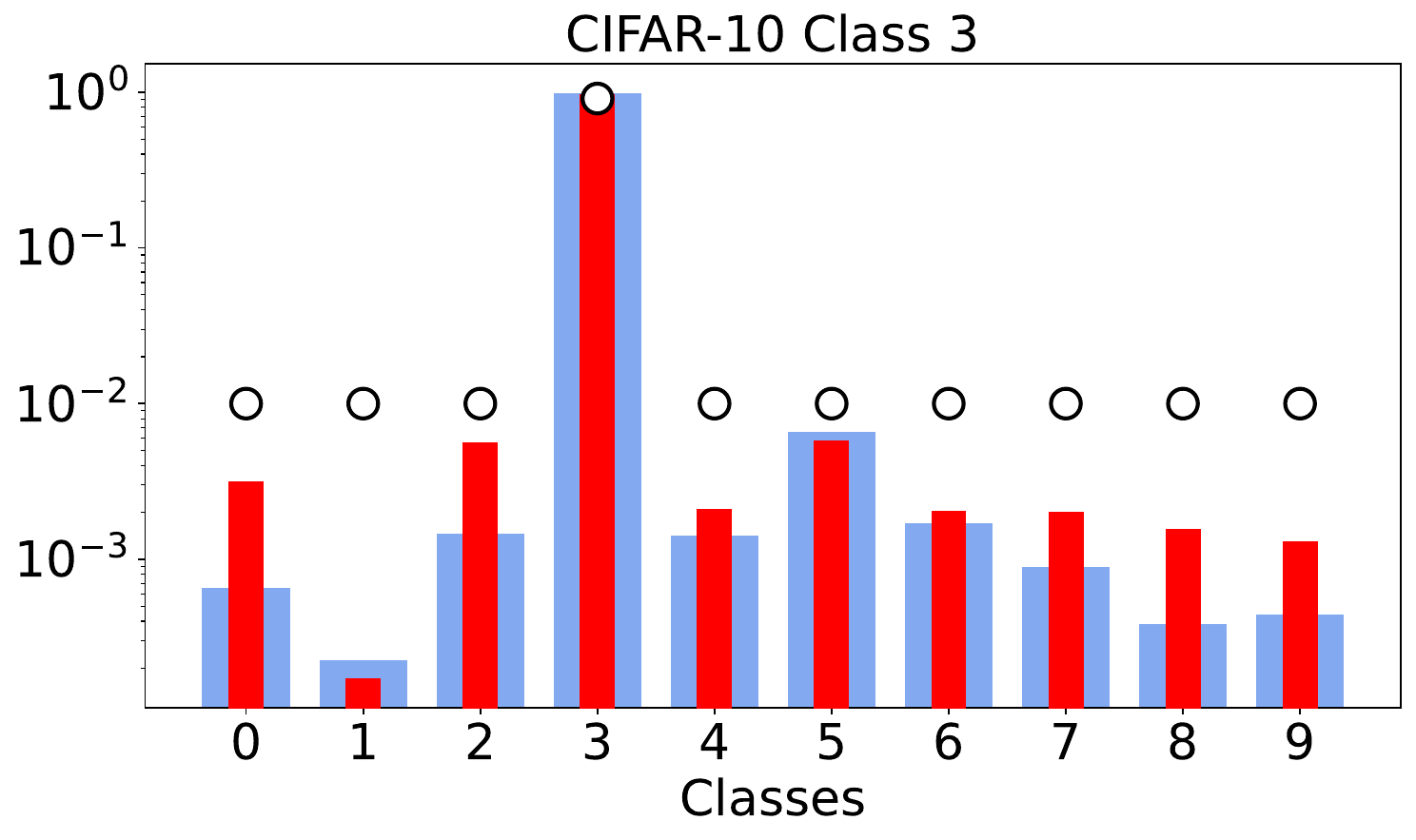}
	\end{subfigure}
	\begin{subfigure}{0.33\linewidth}
		\centering
		\includegraphics[width=\linewidth]{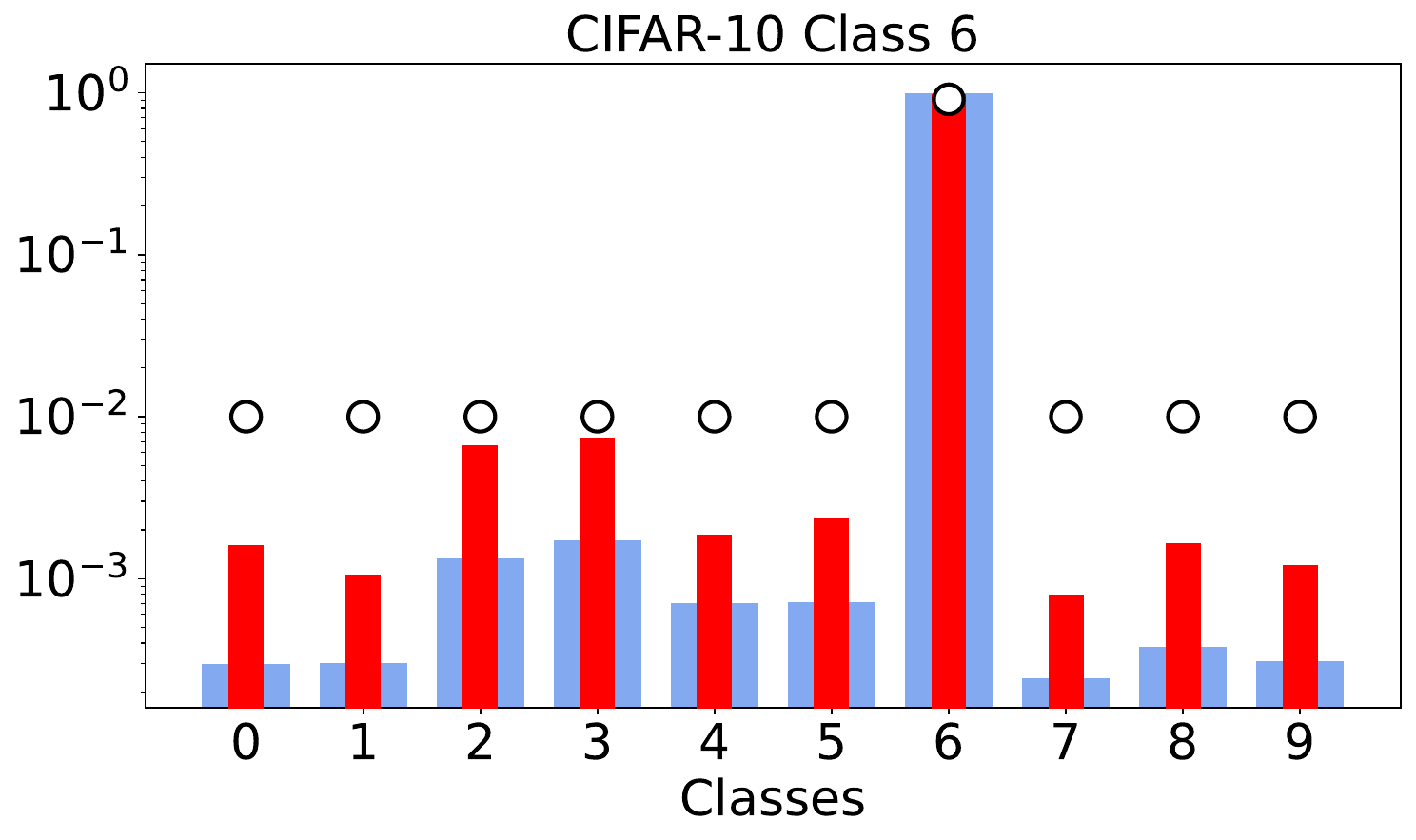}
	\end{subfigure} \\
	\vspace{0.3cm}
	\begin{subfigure}{0.33\linewidth}
		\centering
		\includegraphics[width=\linewidth]{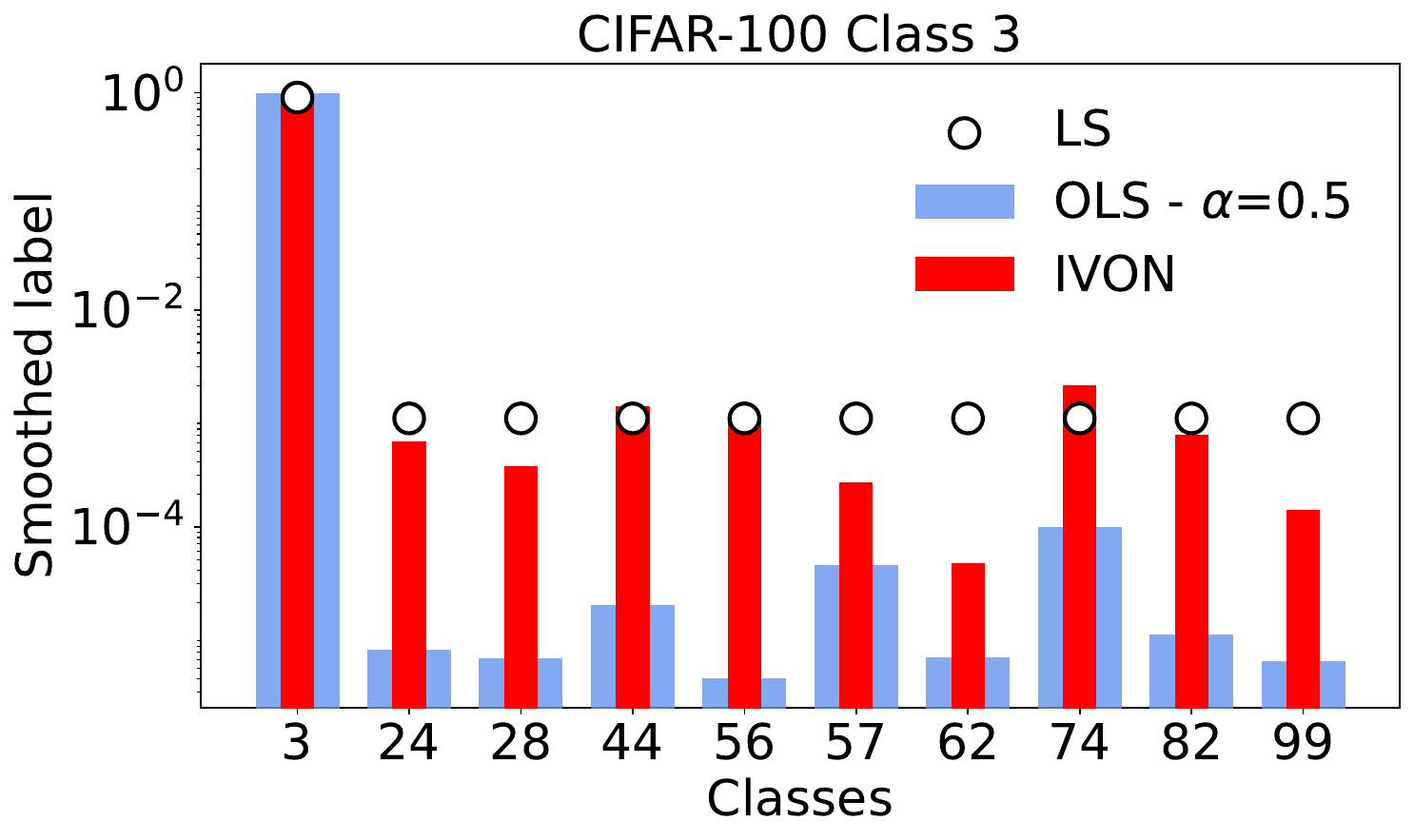}
	\end{subfigure}
	\begin{subfigure}{0.33\linewidth}
		\centering
		\includegraphics[width=\linewidth]{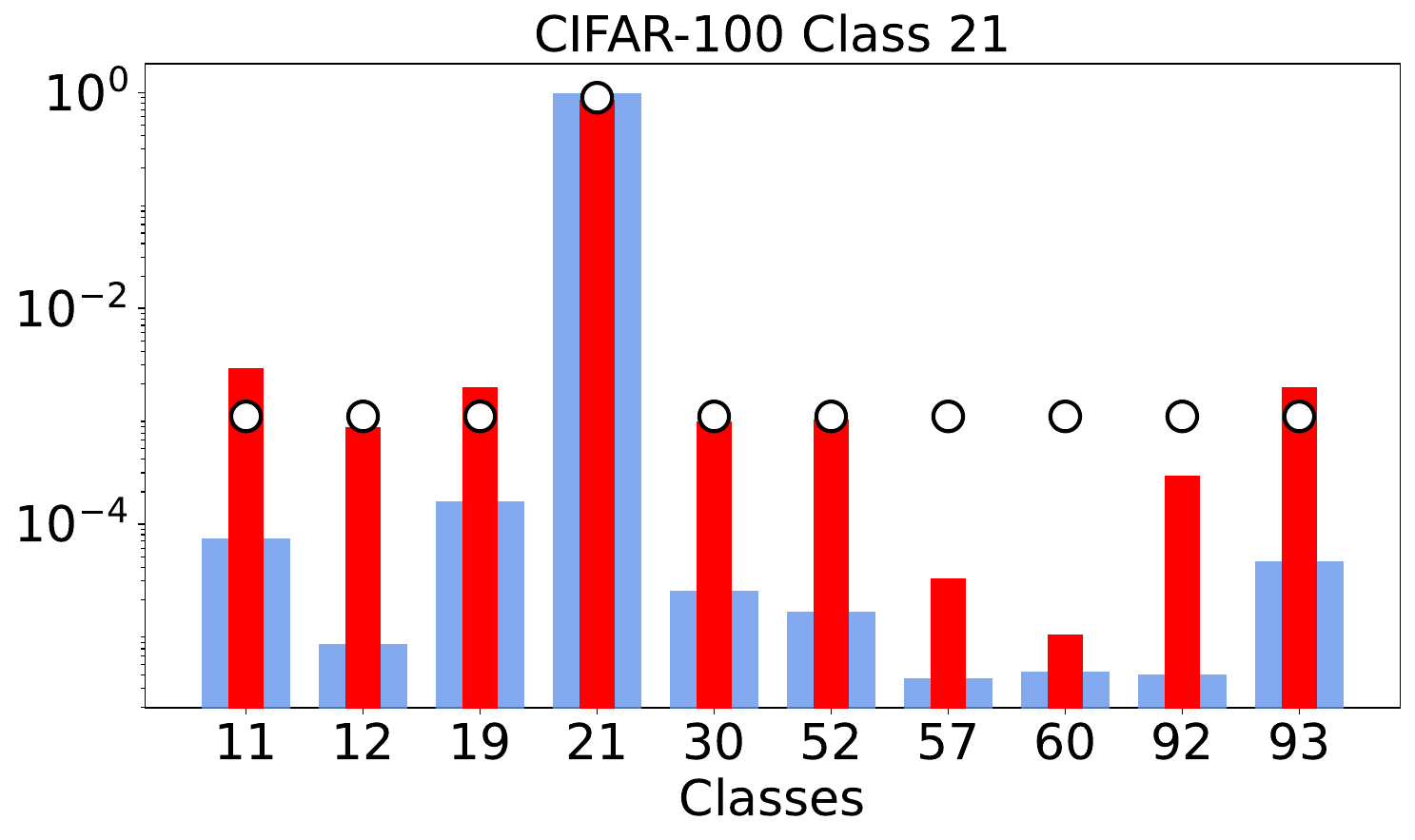}
	\end{subfigure}
	\begin{subfigure}{0.33\linewidth}
		\centering
		\includegraphics[width=\linewidth]{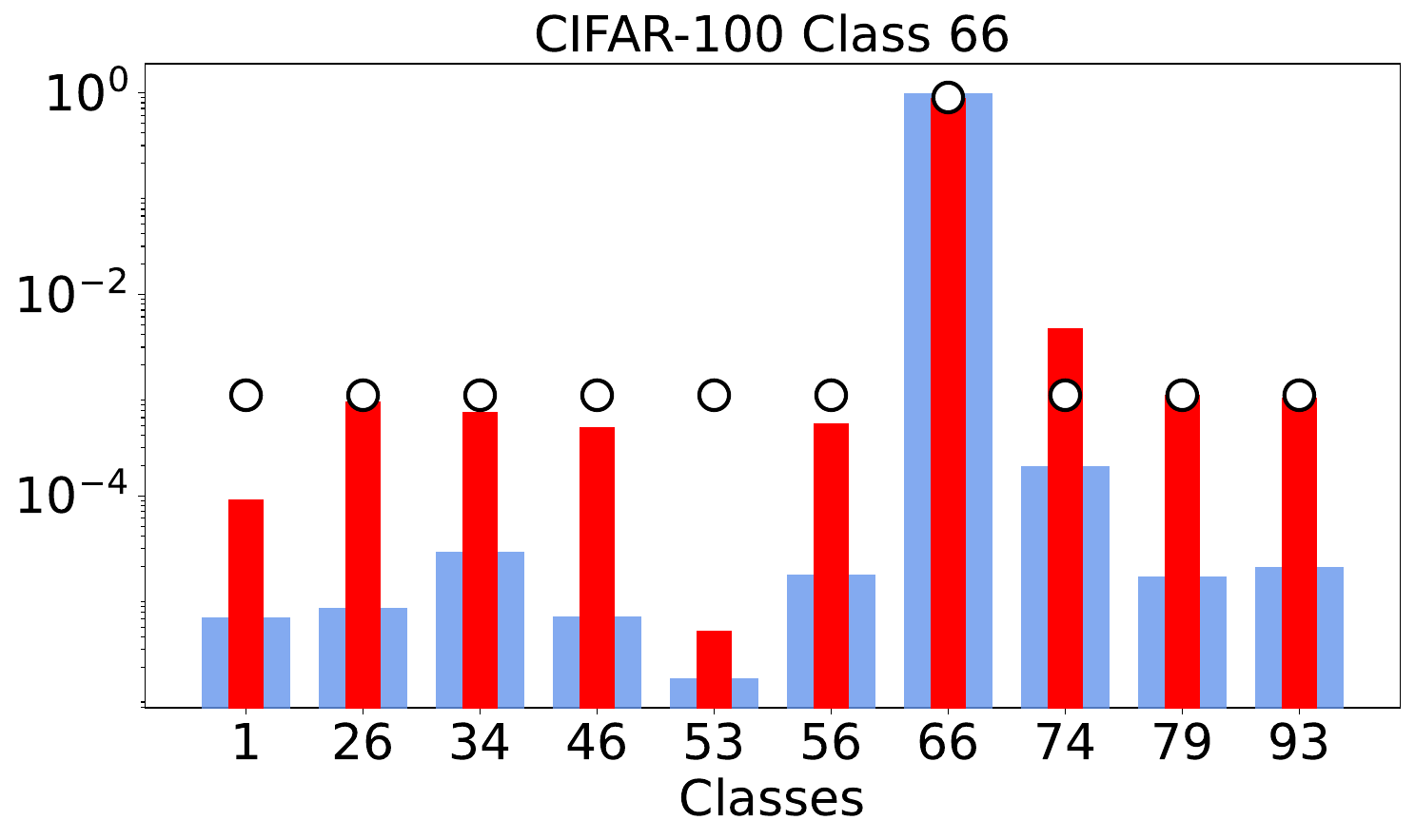}
	\end{subfigure}
	\caption{Smoothed label comparison among IVON \citep{IVON}, LS \citep{Szegedy_rethinking} and Online Label Smoothing (OLS) \citep{zhang2021delving}. IVON has a similar adaptive label smoothing effect as OLS. $\alpha$ is the smoothing rate defined in \cref{equ: label smoothing def}. Y-axis is in the log scale. We randomly pick 10 classes for CIFAR-100 due to image size limit.}
	\label{fig: ols_comparison}
\end{figure*}

We do extensive experiments to show adaptive label noise via variational learning and its benefits. In \cref{sec: noise analyses}, we show that IVON adapts the label noise for each example, and generally assigns higher noise magnitude to ambiguous ones. In \cref{sec: als comparison}, we show that IVON's smoothed labels are similar to an existing adaptive smoothing method \citep{zhang2021delving}. In \cref{sec: mislabelled data}, we show that IVON consistently outperforms LS when datasets have labeling errors in various settings. Additional experiments are reported in \cref{sec: additional exp}, and experiment details are reported in \cref{sec: exp details}.

\begin{figure*}[t!]
	\centering
	\includegraphics[width=0.83\textwidth]{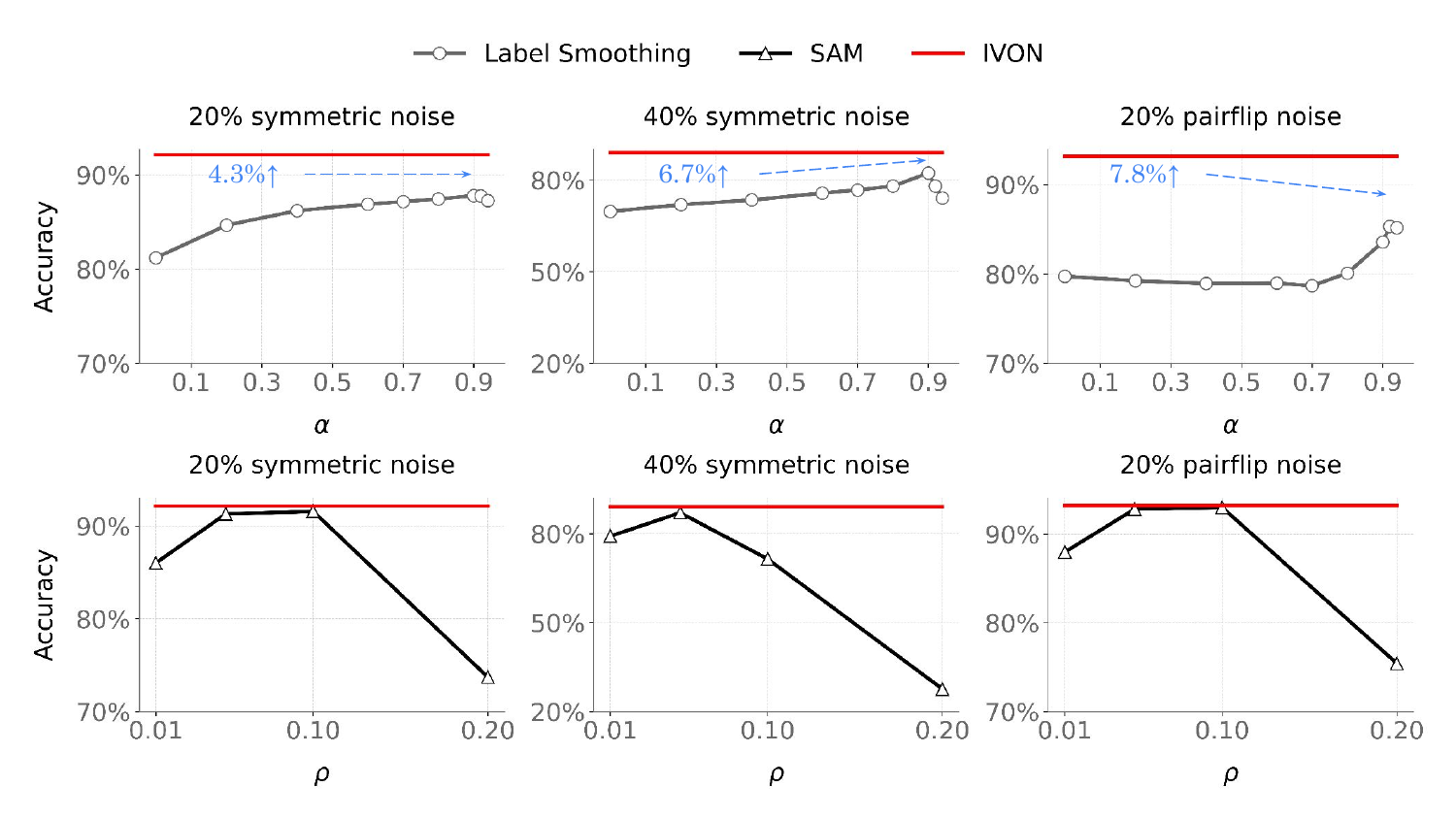}
	\caption{Results on CIFAR-10 with symmetric noisy labels. Top: IVON outperforms Label Smoothing (LS) with different smoothing rates $\alpha$. Down: IVON has comparable results with SAM peak performances, while SAM is sensitive to the choice of perturbation $\rho$. Accuracy improvements are shown in blue. Results are reported over 5 random seeds.}
	\label{fig:cifar10_all}
\end{figure*}

\subsection{IVON's Adaptive Label Noise}\label{sec: noise analyses}

We demonstrate IVON label noise's adaptivity on MNIST dataset \citep{lecun-mnisthandwrittendigit-2010}. We plot IVON's label noise distribution in \cref{fig: mnist_noise_distribution}, which shows that IVON adds different label noise on each example whereas traditional Label Smoothing defines a uniformly distributed noise for all. By further visualizing the data, we see that IVON induces stronger noise to unclear examples, which prevent models from being overconfident in these datapoints. 

\subsection{Comparisons to Existing Adaptive LS Strategies}\label{sec: als comparison}

In this section, we show that IVON's label smoothing is similar to an adaptive method called Online Label Smoothing (OLS) \citep{zhang2021delving}. In the CIFAR-10 and CIFAR-100 dataset \citep{krizhevsky2009cifar}, we compare the smoothed labels of IVON with traditional LS \citep{Szegedy_rethinking} and Online Label Smoothing (OLS) \citep{zhang2021delving}. OLS adjusts the label noise according to the model's predictions, as described in \cref{sec: label smoothing}. As \cref{fig: ols_comparison} shows, IVON has surprisingly similar smoothed label distributions as the OLS in both datasets, while IVON tends to induce stronger label noises. Variational learning's adaptive label smoothing is similar to existing work's, without needing any additional effort to design or estimate the adaptive label noise.

\subsection{Comparisons on Datasets with Labeling Errors} \label{sec: mislabelled data}

 We compare IVON to Label Smoothing (LS) \citep{Szegedy_rethinking} and SAM \citep{SAM_org} in presence of labeling errors, and the results show that IVON consistently outperforms LS in various settings. To find the best performance of the baselines, we tune several LS's smoothing rates $\alpha$ (defined in \cref{equ: label smoothing def}), and various SAM's adversarial perturbation size $\rho$ (discussed in \cref{sec: IVON on NN}). We conduct studies on benchmark datasets with synthetic noise, where the noise level can be adjusted, followed by evaluations on datasets with natural noise, where the noise level is fixed and unknown. For synthetic noise experiments, we use the CIFAR-10 and CIFAR-100 datasets \citep{yu2019does}. For natural noise experiments, we use the benchmark Clothing1M \citep{xiao2015learning}. All datasets include a clean test set. 

\subsubsection{Synthetic Noisy Datasets} \label{sec: synthetic cifar}
We consider two commonly used corruptions \citep{patrini2017making, li2019learning, yu2019does}: Symmetric flipping and Pair flipping. In symmetric flipping, a true label is replaced by a randomly generated class with a probability. In pair flipping, it tries to mimic real world mistakes for similar classes, where a true label is replaced by the next class with a probability. For training dataset, we use previous work's  \citep{yu2019does} code to generate noisy labels. More experiment details are in \cref{sec: cifar_exp_details}.

\begin{figure*}[t!]
	\centering
	\includegraphics[width=0.83\textwidth]{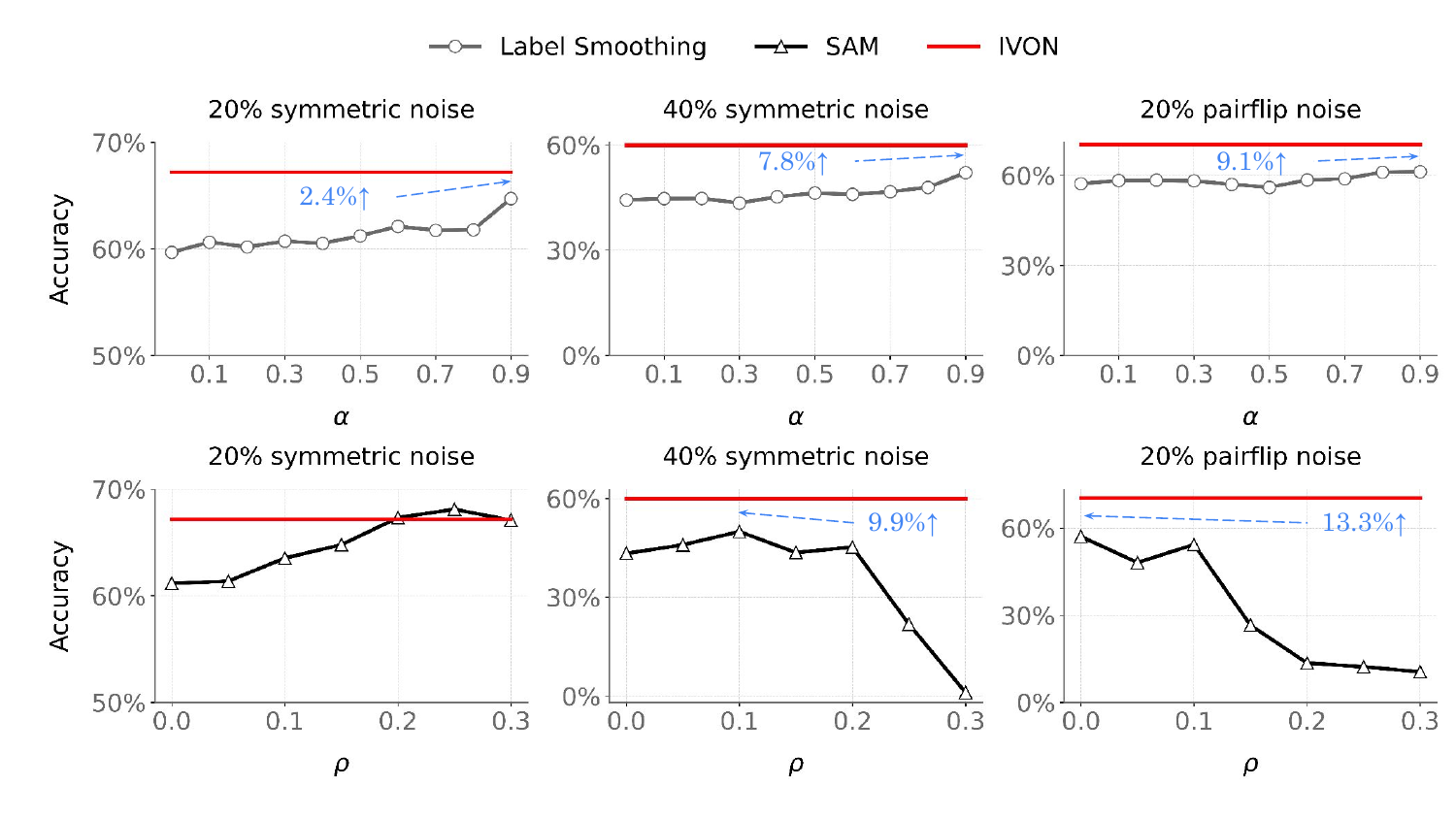}
	\caption{Results on CIFAR-100 with symmetric noisy labels over 5 random seeds, which is similar to CIFAR-10 results in \cref{fig:cifar10_all}.}
	\label{fig:cifar100_all}
\end{figure*}

In CIFAR-10, \cref{fig:cifar10_all} shows that IVON outperforms Label Smoothing and SAM in different scenarios. We also observe that SAM is sensitive to the choice of $\rho$, while IVON does not need to tune any hyperparameters to perform well. In CIFAR-100, \cref{fig:cifar100_all} shows similar trends. For instance, in pairflip 20\% noise setting, IVON outperforms best LS performance by 9.1\% and best SAM performance by 13.3\%. 

\begin{figure}[t]
    \centering 
    \includegraphics[width=\linewidth]{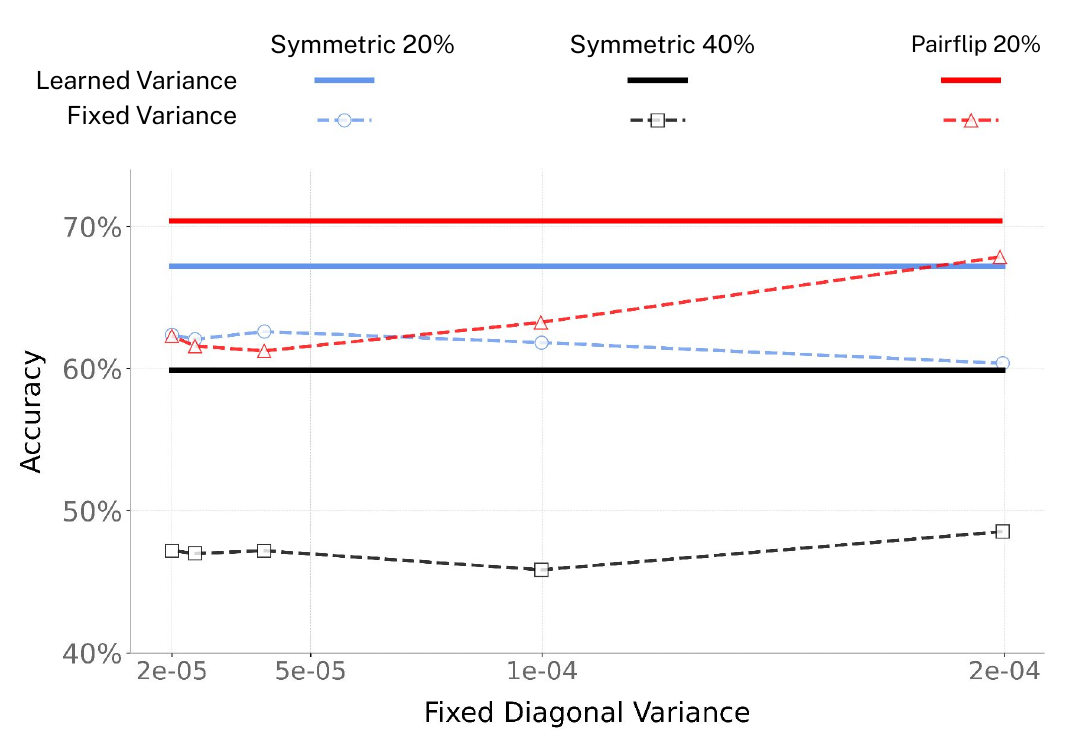}
    \caption{In synthetic noisy datasets of CIFAR-100, we test IVON with multiple fix diagonal variance $\vSigma_t$. The fixed diagonal $\vSigma_t$ is worse than learned diagonal variance in all datasets.}
        \label{fig:cifar100_fix_cov}
\end{figure}

Meanwhile, we test the effectiveness of flexible $\vSigma_t$ by comparing it with the fixed diagonal variance. \cref{fig:cifar100_fix_cov} shows that learned $\vSigma_t$ consistently outperforms fixed $\vSigma_t$ in three noisy datasets. The experiment results demonstrate the importance of flexible $\vSigma_t$ as stated in \cref{sec: newton on GLM}.

\subsubsection{Data Dependent Labeling Errors} \label{sec: data dependent noise}
In this experiment, we try to understand the adaptivity of these methods in the data-dependent noisy dataset. When each class has different noise levels, we expect LS will fail since it adds uniform noises to all classes, while IVON's adaptivity makes it stand.

\begin{figure}[t]
    \centering
    \includegraphics[height=5.6cm]{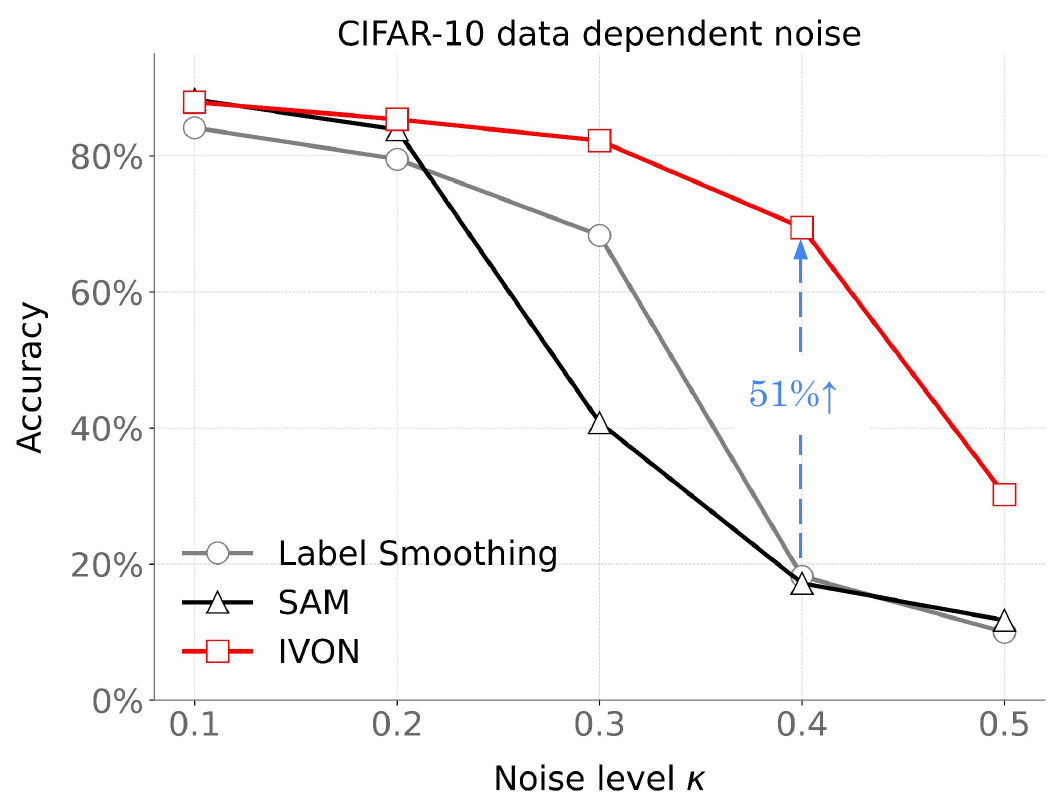}
    \caption{Results for CIFAR-10 with data dependent noise. IVON outperforms LS and SAM in all noise levels. Furthermore, IVON can learn extremely noisy scenario, while LS and SAM cannot.}
    \label{fig:cifar10_inc_acc}
\end{figure}

First, we create a new transition matrix $P$ of noisy label $\mathbf{y}'=P\mathbf{y}$, where $\mathbf{y, y'} \in \mathbb{R}^K, P \in \mathbb{R}^{K\times K}$. We inject difference noise level to each class, so the noise level of each class is different:
\begin{equation} \label{equ: diagonal noise}
    P_{i,i}=1-(\kappa+\beta i), i \in [1,K].
\end{equation}
where $\kappa$ is the starting noise level and $\beta$ is the increase factor. Afterwards, we give the same transition probability to the rest of the wrong classes:
\begin{equation}
    P_{i,j}=\frac{\kappa+\beta i}{K-1}, i,j \in [1,K], i\neq j.
\end{equation}

In experiments, we follow the hyperparameters in CIFAR-10 synthetic noise experiment from \cref{sec: synthetic cifar}. For LS, we run smoothing rate $\{0,0.1,0.3,0.5,0.7,0.9\}$ and report the best accuracy. For SAM, we run $\rho$ for $\{0,0.05,0.1,0.15,0.2,0.5\}$ and report the best accuracy.

The experiment results for $\kappa= \{0.1 \sim 0.5\}$ and $\beta=0.05$ are in \cref{fig:cifar10_inc_acc}. Overall, IVON outperforms LS and SAM in all noise levels. Meanwhile, IVON can learn in very noisy scenarios $\kappa=\{0.4, 0.5\}$ while baselines can only reach around $10\%$ accuracy. The experiment results support our claim that adaptive label noise induced by variational learning is more effective than traditional label smoothing.

\subsubsection{Uncontrolled Noisy Datasets}
We now report results on Clothing1M \citep{xiao2015learning}, a large-scale dataset that features natural label noise from the web and consists of 1 million images across 14 categories. We conduct experiments by using ResNet-50 as the model. 

\begin{figure}[h]
	\centering
	\includegraphics[width=0.92\linewidth]{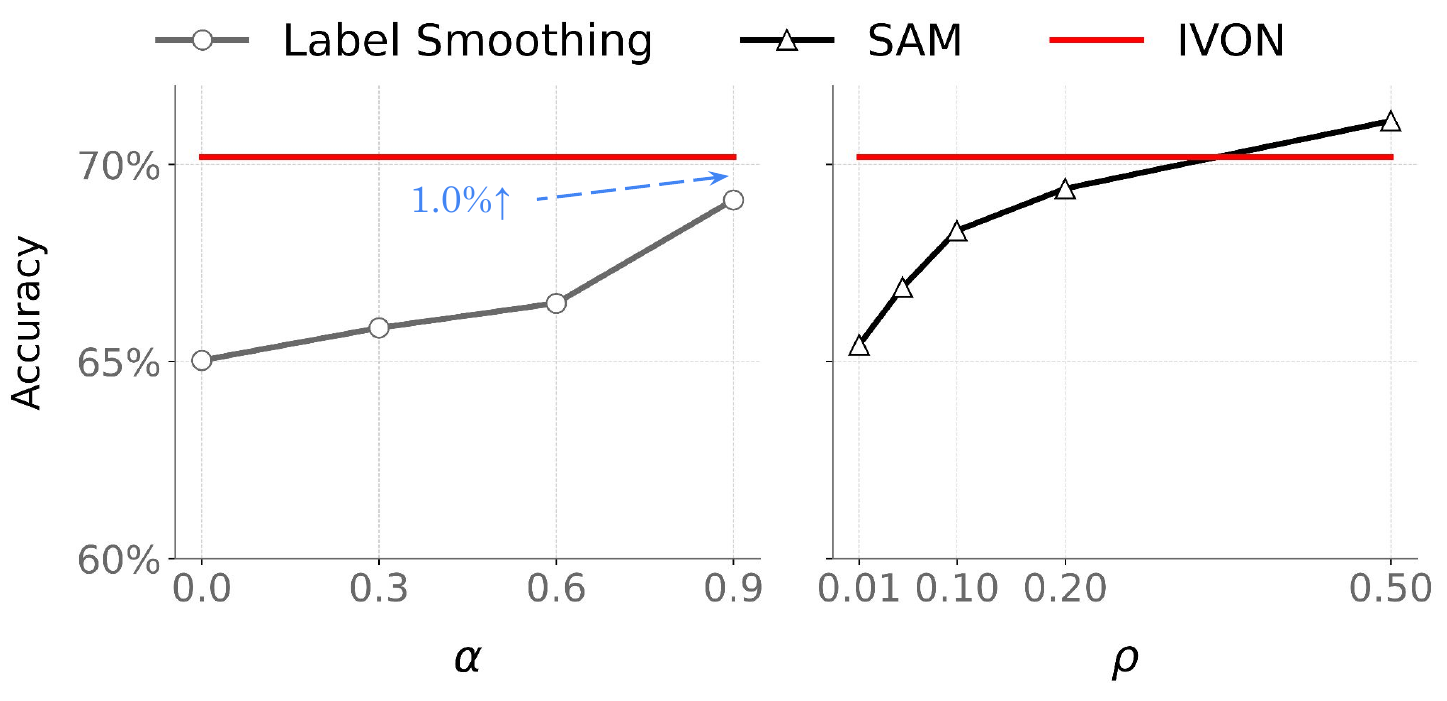}
	\caption{ Clothing 1M experiment result. The result is similar to synthetic noisy datasets reported in \cref{fig:cifar10_all} and \cref{fig:cifar100_all}. Results are reported over 5 seeds.}
	\label{fig: clothing1m_all}
\end{figure}

The results on Clothing1M, illustrated in \cref{fig: clothing1m_all}, demonstrate that IVON outperforms Label Smoothing and is comparable to SAM. This experiment shows that IVON's performance is consistent in the large scale dataset.

 \section{Conclusion}
 In this paper, we show that variational learning induces an adaptive label smoothing similar to an existing adaptive approach \citep{zhang2021delving} but does not require any additional effort to design. We derive the exact form for simple models and extend them to neural networks. We empirically confirm the effectiveness of noise, showing that the IVON method consistently performs better than LS, and comparably to SAM, without requiring hyperparameters to achieve desired smoothing. Our work suggests that Bayesian frameworks are naturally suitable for label noise. Specifically, we believe that variational learning algorithms, such as IVON, provide a flexible framework to further add noise to handle both the abnormalities and typicalities in the data.

\clearpage

\section*{Impact Statement}
This paper presents work whose goal is to advance the field of 
Machine Learning. There are many potential societal consequences 
of our work, none which we feel must be specifically highlighted here. 

\section*{Acknowledgements}
Mohammad Emtiyaz Khan was supported by the Bayes duality project, JST CREST Grant Number JPMJCR2112. Some of the experiments were carried out using the TSUBAME4.0 supercomputer at Institute of Science Tokyo.

We also would like to thank our former post-doc Lu Xu who started this project and contributed heavily to it. For reasons beyond our control, Lu was unable to join the author list but this work would not have been possible without her work. We also would like to thank Thomas Möllenhoff (RIKEN) for his help in deriving some of the results.

\section*{Author Contributions Statement}
Author list: Sin-Han Yang (SHY), Zhedong Liu (ZL), Gian Maria Marconi (GMM) and Mohammad Emtiyaz Khan (MEK).

MEK proposed the label-noise idea which was developed by Lu Xu and GMM who also performed the first set of experiments to validate the initial hypothesis.  Subsequently, ZL extended these experiments to the Clothing 1M settings.  Building up on all these experiments, SHY did most of the experiments involving IVON which are currently in the final version of the paper. MEK wrote the theory in Sec 3 with help from SHY. Both SHY and MEK contributed to the final writing with feedback from ZL and GMM.

\bibliography{reference}

\begin{thebibliography}{27}
\providecommand{\natexlab}[1]{#1}
\providecommand{\url}[1]{\texttt{#1}}
\expandafter\ifx\csname urlstyle\endcsname\relax
  \providecommand{\doi}[1]{doi: #1}\else
  \providecommand{\doi}{doi: \begingroup \urlstyle{rm}\Url}\fi

\bibitem[Damian et~al.(2021)Damian, Ma, and Lee]{damian2021label}
Damian, A., Ma, T., and Lee, J.~D.
\newblock Label noise {SGD} provably prefers flat global minimizers.
\newblock \emph{Advances in Neural Information Processing Systems}, 34:\penalty0 27449--27461, 2021.

\bibitem[Foret et~al.(2021)Foret, Kleiner, Mobahi, and Neyshabur]{SAM_org}
Foret, P., Kleiner, A., Mobahi, H., and Neyshabur, B.
\newblock Sharpness-aware minimization for efficiently improving generalization.
\newblock In \emph{International Conference on Learning Representations, {ICLR} 2021, Virtual Event, Austria, May 3-7, 2021}, 2021.

\bibitem[Ghoshal et~al.(2021)Ghoshal, Chen, Gupta, Zettlemoyer, and Mehdad]{ghoshal2021learning}
Ghoshal, A., Chen, X., Gupta, S., Zettlemoyer, L., and Mehdad, Y.
\newblock Learning better structured representations using low-rank adaptive label smoothing.
\newblock In \emph{International Conference on Learning Representations}, 2021.

\bibitem[Khan et~al.(2018)Khan, Nielsen, Tangkaratt, Lin, Gal, and Srivastava]{khan2018fastadam}
Khan, M., Nielsen, D., Tangkaratt, V., Lin, W., Gal, Y., and Srivastava, A.
\newblock Fast and scalable {B}ayesian deep learning by weight-perturbation in {A}dam.
\newblock In \emph{International conference on machine learning}, pp.\  2611--2620. PMLR, 2018.

\bibitem[Khan \& Rue(2023)Khan and Rue]{khan2023bayesian}
Khan, M.~E. and Rue, H.
\newblock The {B}ayesian learning rule.
\newblock \emph{Journal of Machine Learning Research}, 24\penalty0 (281):\penalty0 1--46, 2023.

\bibitem[Khan et~al.(2019)Khan, Immer, Abedi, and Korzepa]{khan2019approximate}
Khan, M.~E., Immer, A., Abedi, E., and Korzepa, M.
\newblock Approximate inference turns deep networks into {G}aussian processes.
\newblock \emph{Advances in Neural Information Processing Systems}, 2019.

\bibitem[Ko et~al.(2023)Ko, Yi, and Yun]{ko2023gift}
Ko, J., Yi, B., and Yun, S.-Y.
\newblock A gift from label smoothing: robust training with adaptive label smoothing via auxiliary classifier under label noise.
\newblock In \emph{Proceedings of the AAAI Conference on Artificial Intelligence}, 2023.

\bibitem[Krizhevsky \& Hinton(2009)Krizhevsky and Hinton]{krizhevsky2009cifar}
Krizhevsky, A. and Hinton, G.
\newblock Learning multiple layers of features from tiny images.
\newblock Technical report, Citeseer, 2009.

\bibitem[LeCun \& Cortes(2010)LeCun and Cortes]{lecun-mnisthandwrittendigit-2010}
LeCun, Y. and Cortes, C.
\newblock {MNIST} handwritten digit database.
\newblock 2010.
\newblock URL \url{http://yann.lecun.com/exdb/mnist/}.

\bibitem[Lee et~al.(2022)Lee, Cheung, and Zhang]{lee2022adaptive}
Lee, D., Cheung, K.~C., and Zhang, N.~L.
\newblock Adaptive label smoothing with self-knowledge in natural language generation.
\newblock In Goldberg, Y., Kozareva, Z., and Zhang, Y. (eds.), \emph{Proceedings of the 2022 Conference on Empirical Methods in Natural Language Processing, {EMNLP} 2022}, pp.\  9781--9792. Association for Computational Linguistics, 2022.

\bibitem[Li et~al.(2019)Li, Wong, Zhao, and Kankanhalli]{li2019learning}
Li, J., Wong, Y., Zhao, Q., and Kankanhalli, M.~S.
\newblock Learning to learn from noisy labeled data.
\newblock In \emph{{IEEE} Conference on Computer Vision and Pattern Recognition, {CVPR} 2019}, pp.\  5051--5059. Computer Vision Foundation / {IEEE}, 2019.

\bibitem[Liu(2021)]{liu2021understanding}
Liu, Y.
\newblock Understanding instance-level label noise: Disparate impacts and treatments.
\newblock In \emph{International Conference on Machine Learning}, pp.\  6725--6735. PMLR, 2021.

\bibitem[Lukasik et~al.(2020)Lukasik, Bhojanapalli, Menon, and Kumar]{lukasik2020does}
Lukasik, M., Bhojanapalli, S., Menon, A.~K., and Kumar, S.
\newblock Does label smoothing mitigate label noise?
\newblock In \emph{Proceedings of the 37th International Conference on Machine Learning, {ICML} 2020}, volume 119 of \emph{Proceedings of Machine Learning Research}, pp.\  6448--6458. {PMLR}, 2020.

\bibitem[M{\"{o}}llenhoff \& Khan(2023)M{\"{o}}llenhoff and Khan]{thomassampaper}
M{\"{o}}llenhoff, T. and Khan, M.~E.
\newblock {SAM} as an optimal relaxation of bayes.
\newblock In \emph{The Eleventh International Conference on Learning Representations, {ICLR} 2023}. OpenReview.net, 2023.

\bibitem[M{\"u}ller et~al.(2019)M{\"u}ller, Kornblith, and Hinton]{muller2019does}
M{\"u}ller, R., Kornblith, S., and Hinton, G.~E.
\newblock When does label smoothing help?
\newblock \emph{Advances in neural information processing systems}, 32, 2019.

\bibitem[Nickl et~al.(2024)Nickl, Xu, Tailor, M{\"o}llenhoff, and Khan]{nickl2024memory}
Nickl, P., Xu, L., Tailor, D., M{\"o}llenhoff, T., and Khan, M.~E.
\newblock The memory-perturbation equation: Understanding model's sensitivity to data.
\newblock \emph{Advances in Neural Information Processing Systems}, 36, 2024.

\bibitem[Park et~al.(2023)Park, Noh, Oh, Baek, and Ham]{park2023acls}
Park, H., Noh, J., Oh, Y., Baek, D., and Ham, B.
\newblock {ACLS}: Adaptive and conditional label smoothing for network calibration.
\newblock In \emph{Proceedings of the IEEE/CVF International Conference on Computer Vision}, pp.\  3936--3945, 2023.

\bibitem[Patrini et~al.(2017)Patrini, Rozza, Menon, Nock, and Qu]{patrini2017making}
Patrini, G., Rozza, A., Menon, A.~K., Nock, R., and Qu, L.
\newblock Making deep neural networks robust to label noise: {A} loss correction approach.
\newblock In \emph{2017 {IEEE} Conference on Computer Vision and Pattern Recognition, {CVPR} 2017}, pp.\  2233--2241. {IEEE} Computer Society, 2017.

\bibitem[Pereyra et~al.(2017{\natexlab{a}})Pereyra, Tucker, Chorowski, Kaiser, and Hinton]{pereyra2017regularizing}
Pereyra, G., Tucker, G., Chorowski, J., Kaiser, {\L}., and Hinton, G.
\newblock Regularizing neural networks by penalizing confident output distributions.
\newblock \emph{arXiv preprint arXiv:1701.06548}, 2017{\natexlab{a}}.

\bibitem[Pereyra et~al.(2017{\natexlab{b}})Pereyra, Tucker, Chorowski, Kaiser, and Hinton]{li2020regularization}
Pereyra, G., Tucker, G., Chorowski, J., Kaiser, L., and Hinton, G.~E.
\newblock Regularizing neural networks by penalizing confident output distributions.
\newblock In \emph{5th International Conference on Learning Representations, {ICLR} 2017, Toulon, France, April 24-26, 2017, Workshop Track Proceedings}. OpenReview.net, 2017{\natexlab{b}}.

\bibitem[Shen et~al.(2024)Shen, Daheim, Cong, Nickl, Marconi, Bazan, Yokota, Gurevych, Cremers, Khan, and M{\"{o}}llenhoff]{IVON}
Shen, Y., Daheim, N., Cong, B., Nickl, P., Marconi, G.~M., Bazan, C., Yokota, R., Gurevych, I., Cremers, D., Khan, M.~E., and M{\"{o}}llenhoff, T.
\newblock Variational learning is effective for large deep networks.
\newblock In \emph{Forty-first International Conference on Machine Learning, {ICML} 2024}. OpenReview.net, 2024.

\bibitem[Szegedy et~al.(2016)Szegedy, Vanhoucke, Ioffe, Shlens, and Wojna]{Szegedy_rethinking}
Szegedy, C., Vanhoucke, V., Ioffe, S., Shlens, J., and Wojna, Z.
\newblock Rethinking the inception architecture for computer vision.
\newblock In \emph{2016 IEEE Conference on Computer Vision and Pattern Recognition (CVPR)}, pp.\  2818--2826, 2016.

\bibitem[Xiao et~al.(2015)Xiao, Xia, Yang, Huang, and Wang]{xiao2015learning}
Xiao, T., Xia, T., Yang, Y., Huang, C., and Wang, X.
\newblock Learning from massive noisy labeled data for image classification.
\newblock In \emph{Proceedings of the IEEE Conference on Computer Vision and Pattern Recognition}, 2015.

\bibitem[Xu et~al.(2024)Xu, Lee, Yoon, and Park]{xu2024adaptive}
Xu, M., Lee, J., Yoon, S., and Park, D.~S.
\newblock Adaptive label smoothing for out-of-distribution detection.
\newblock \emph{arXiv preprint arXiv:2410.06134}, 2024.

\bibitem[Yu et~al.(2019)Yu, Han, Yao, Niu, Tsang, and Sugiyama]{yu2019does}
Yu, X., Han, B., Yao, J., Niu, G., Tsang, I.~W., and Sugiyama, M.
\newblock How does disagreement help generalization against label corruption?
\newblock In Chaudhuri, K. and Salakhutdinov, R. (eds.), \emph{Proceedings of the 36th International Conference on Machine Learning, {ICML} 2019}, volume~97, pp.\  7164--7173. {PMLR}, 2019.

\bibitem[Zhang et~al.(2021)Zhang, Jiang, Hou, Wei, Han, Li, and Cheng]{zhang2021delving}
Zhang, C.-B., Jiang, P.-T., Hou, Q., Wei, Y., Han, Q., Li, Z., and Cheng, M.-M.
\newblock Delving deep into label smoothing.
\newblock \emph{IEEE Transactions on Image Processing}, 30:\penalty0 5984--5996, 2021.

\bibitem[Zhang et~al.(2018)Zhang, Sun, Duvenaud, and Grosse]{zhang2018noisy}
Zhang, G., Sun, S., Duvenaud, D., and Grosse, R.
\newblock Noisy natural gradient as variational inference.
\newblock In \emph{International conference on machine learning}, pp.\  5852--5861. PMLR, 2018.

\end{thebibliography}
\bibliographystyle{icml2025}

\newpage
\appendix
\onecolumn
\section{Derivations}

\subsection{Derivation of GLM with Newton's method}
\label{app:glm_newton}
Newton's update shown in \cref{eq:newton} is equivalent to the following surrogate minimization,
 \begin{equation}
       \vparam_{t+1} = \,\, \arg\min_{\vparam} \,\,  \vparam^\top \grad \barloss(\vparam_t) + \frac{1}{2} (\vparam-\vparam_t)^\top \nabla^2 \barloss(\vparam_t) (\vparam-\vparam_t) .
\label{eq:newton_surrogate}
\end{equation}
This can be verified by simply taking the derivative of the above objective and setting it to zero. We will now show that the VON update give rise to a similar surrogate but where the gradient and Hessian are replaced by their expected values. 

To do so, we use the result of \citet{khan2019approximate} who show that each step of VON algorithm can be seen as inference on a linear model. Essentially, the VON update can be expressed as follows (see \citet[App. C.3]{nickl2024memory} for a derivation):
 \begin{equation}
    \begin{split}
       q_{t+1}(\vparam) &\propto q_t(\vparam)^{1-\rho_t} \prod_{i=1}^N e^{ \rho_t \rnd{ \vparam^\top \myexpect_{q_t}[-\nabla \loss_i(\vparam) + \nabla^2 \loss_i(\vparam) \vparam_t] -\half \vparam^\top \myexpect_{q_t}[\nabla^2 \loss_i(\vparam)] \vparam} } \\
       &\propto q_t(\vparam)^{1-\rho_t} \prod_{i=1}^N e^{ -\rho_t \rnd{ \vparam^\top \myexpect_{q_t}[\nabla \loss_i(\vparam)] + \half (\vparam - \vparam_t)^\top \myexpect_{q_t}[\nabla^2 \loss_i(\vparam)] (\vparam-\vparam_t)} }, 
    \end{split}
    \label{eq:blr_sgd}
 \end{equation}
 where we subtracted $\vparam_t^\top \myexpect_{q_t}[\nabla^2 \loss_i(\vparam)] \vparam_t$ and completed the square. This is a constant which is absorbed in the normalizing constant of $q_{t+1}$. From here, we can simply match the mode $\vparam_{t+1}$ of $q_{t+1}$ to the mode of the right hand side. For $\rho_t = 1$, this gives us the following minimization problem to recover $\vparam_{t+1}$:
\begin{equation}
   \vparam_{t+1} = \,\, \arg\min_{\vparam} \,\,  \vparam^\top \myexpect_{q_t}[\grad \barloss(\vparam)] + \frac{1}{2} (\vparam-\vparam_t)^\top \myexpect_{q_t}[\nabla^2 \barloss(\vparam)] (\vparam-\vparam_t).
\end{equation}
This shows that, for $\rho_t =1$, VON updates can be seen as Newton step where gradient and Hessian are replaced by their expected values. The proof is identical to the one shown in the main tex, therefore we omit it.

 \begin{thm}
    For the loss function of \cref{eq:bce}, the VON update in \cref{eq:von} with $\rho_t = 1$ is equivalent to Newton's update in \cref{eq:newton} but where the label $y_i$ are replaced by $y_i + \epsilon_{i|t}$ with noise defined as
    \begin{equation}
       \epsilon_{i|t} =  \sigmoid\rnd{f_{i|t}} - \myexpect_{\text{\gauss}(e|0,1)} \sqr{\sigmoid\rnd{f_{i|t} + e \sqrt{\vphi_i^\top\vSigma_t \vphi_i}  }},
       \label{eq:sgd_noise}
    \end{equation}
    and the Hessian $\nabla^2 \barloss(\vparam_t)$ is replaced by its noisy version $\myexpect_{q_t}[\nabla^2 \barloss(\vparam)]$.
 \end{thm}
    

\clearpage

\subsection{IVON pseudo code}

The pseudo-code is given in \cref{alg:ivon}.

\definecolor{commentcolor}{RGB}{128, 179, 89}
\renewcommand\algorithmiccomment[1]{\hfill{\textcolor{commentcolor}{\eqparbox{COMMENT}{#1}}}}
\begin{algorithm}[!h]
   \caption{Improved Variational Online Newton (IVON) \citep{IVON}.}
	\label{alg:ivon}
   \begin{algorithmic}[1]
		\setstretch{1.15}
      \REQUIRE Learning rates $\{ \alpha_t \}$, weight-decay $\delta >
      0$.
      \REQUIRE Momentum parameters $\beta_1, \beta_2 \in [0, 1)$.
      \REQUIRE Hessian init $h_0 > 0$.
      \renewcommand{\algorithmicrequire}{\textbf{Init:}}
		\REQUIRE $\vm \leftarrow \text{(NN-weights)}$,\,\, $\vh \leftarrow h_0 $,
                      \,\, $\vg \leftarrow 0$, \,\, $\lambda \leftarrow N$.
      \REQUIRE $\vsigma
                      \leftarrow 1 / \sqrt{\lambda (\vh +
                        \delta)}$.
      \renewcommand{\algorithmicrequire}{\textbf{Optional:}}
      \REQUIRE $\alpha_t \leftarrow (h_0 +
                      \delta) \alpha_t$\,  for all $t$.
                      \FOR{$t=1,2,\hdots$}
                      \STATE \hspace{-0.15cm}$\widehat \vg \leftarrow
                      {\widehat \nabla} \barloss(\vparam)$,
      \text{\textcolor{black}{ where} } 
      $\vparam \sim q$
      \STATE \hspace{-0.15cm}$\widehat \vh \leftarrow {\widehat \vg\cdot (\vparam-\vm) / \vsigma^2}$
		\STATE \hspace{-0.15cm}$\vg \leftarrow \beta_1 \vg\hspace{-0.03cm}+\hspace{-0.05cm}(1\hspace{-0.05cm}-\hspace{-0.05cm}\beta_1) \widehat \vg$ 
                \STATE \hspace{-0.15cm}$\vh \leftarrow \beta_2 \vh+(1
                - \beta_2)\widehat \vh$${+\half (1 - \beta_2)^2
                  (\vh - \widehat \vh)^2 / (\vh + \delta)}$
      \STATE \hspace{-0.15cm}$\bar \vg \leftarrow \vg / (1 - \beta_1^{t})$ 
      \STATE \hspace{-0.15cm}$\vm \leftarrow \vm - \alpha_t(\bar \vg + \delta \vm) / ({\vh} + \delta)$
		\STATE \hspace{-0.15cm}$\vsigma \leftarrow 1 / \sqrt{\lambda (\vh + \delta)}$
      \ENDFOR
		\STATE \textbf{return} $\vm, \vsigma$ 
	\end{algorithmic}
	\setstretch{1}
\end{algorithm}

\section{Additional Experiments} \label{sec: additional exp}
\subsection{Hessian Initialization}

\begin{figure}[h]
    \centering 
    \includegraphics[width=0.61\textwidth]{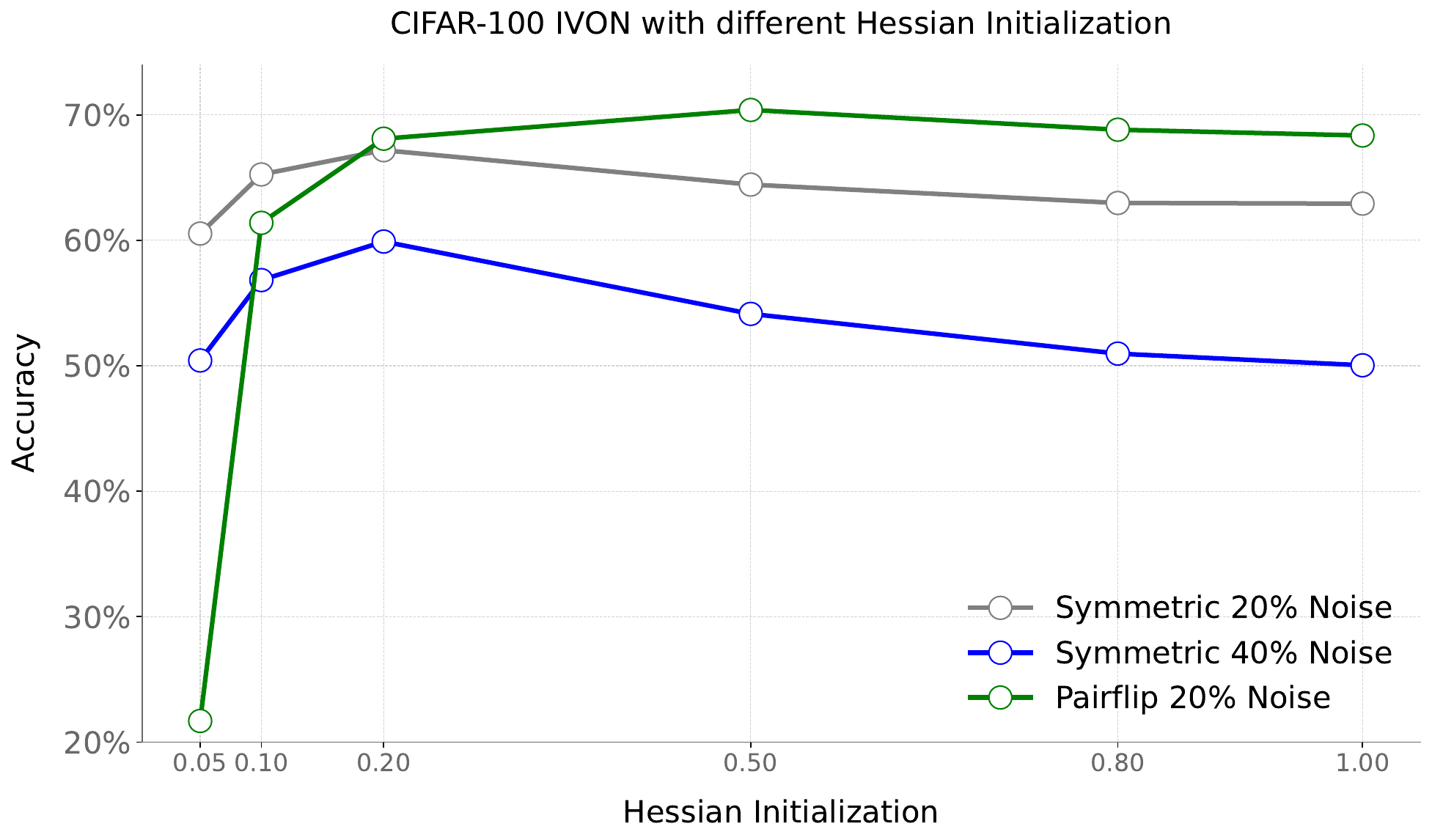}
    \caption{Results for IVON on CIFAR-100 with multiple Hessian initializations. IVON's accuracy is consistent when having different Hessian initializations.}
        \label{fig:cifar100_ivon_hess}
\end{figure}

We analyze how Hessian initialization $h_0$ of IVON affects the accuracy. The results are in \cref{fig:cifar100_ivon_hess}. IVON's accuracy can only vary by up to 10\% when the Hessian is bigger than $0.05$, and this variation is less sensitive compared to SAM's sensitivity to $\rho$, as shown in \cref{fig:cifar10_all}, \cref{fig:cifar100_all} and \cref{fig: clothing1m_all}. 

\section{Experiment details} \label{sec: exp details}
\subsection{Experiment details of \cref{sec: noise analyses} and \cref{sec: als comparison}}

In \cref{fig: mnist_noise_distribution}, we test IVON on a 3-layers convolutional neural networks. In \cref{fig: ols_comparison}, we do experiments on ResNet-34 model. We uses the PyTorch implementation verison\footnote{\href{https://github.com/ankandrew/online-label-smoothing-pt}{https://github.com/ankandrew/online-label-smoothing-pt}} of Online Label Smoothing \citep{zhang2021delving}. 

\subsection{Experiments on Synthetic Noisy Datasets} \label{sec: cifar_exp_details}
For pairflip setting in CIFAR-10, the classes flipping order is: AIRPLANE $\rightarrow$ AUTOMOBILE $\rightarrow$ BIRD $\rightarrow$ CAT $\rightarrow$ DEER $\rightarrow$ DOG $\rightarrow$ FROG $\rightarrow$ HORSE $\rightarrow$ SHIP $\rightarrow$ TRUCK $\rightarrow$ AIRPLANE.
In CIFAR-10 experiments, we train a ResNet 34 for 200 epochs with batch size set to 50 and weight decay set to 0.001. For SAM and LS, we set initial learning rate as $0.05$ and reduce it by 0.1 at 100 epoch and 150 epoch, following hyper-parameters from previous papers. For IVON \citep{IVON}, we follow the original paper to set initial learning rate as $0.2$ and anneal the learning rate to zero with a cosine schedule after a linear warmup phase over 5 epochs. We set momentum to $0.9$ for all methods, and hessian momentum $\beta_2$ to $1 - e^{-5}$, hessian initial $h_0$ to $0.9$, scaling parameter $\lambda$ to the number of training data for IVON. For SAM, we follow the original paper \citep{SAM_org} and choose best neighborhood size $\rho$ from $[0.01, 0.05, 0.1, 0.2, 0.5]$. In CIFAR-100 experiments, we tune the hyperparamters to the best for each method. The hyperparameters are specified in \cref{tab: cifar100 hyperparameter tuning results}. 

In \cref{fig:cifar100_fix_cov}, we fix the Hessian of IVON by setting $\beta_2=1$ in Line $5$ of \cref{alg:ivon}. Therefore, standard deviation $\vsigma$ defined in Line $8$ is fixed since Hessian $\vh$ is fixed.

\begin{table}[ht]
\centering
\caption{Hyperparamters of each method for CIFAR-100. We denote learning rate as lr, Hessian Initialization as Hessian init.}
\label{tab: cifar100 hyperparameter tuning results}
\begin{tabular}{lccc}
\toprule
                   & Symmetric 20\% & Symmetric 40\% & Pairflip 20\% \\
                   \midrule
Weight decay       & 2e-4           & 2e-4           & 5e-4          \\
LS \citep{Szegedy_rethinking} lr & 0.1               &     0.1           &    0.1           \\
SAM \citep{SAM_org} lr             &     0.05           &    0.1            &    0.1           \\
IVON \citep{IVON} lr            &      0.8          &     0.8           &      0.5         \\
IVON \citep{IVON} Hessian init  &      0.2          &     0.2           &      0.5         \\\bottomrule
\end{tabular}
\end{table}
\subsection{Clothing 1M Details}
The noisy labels in Clothing1M \citep{xiao2015learning} are derived from the text surrounding the images on the web. In constructing the dataset, noisy labels are assigned to images based on this contextual text.

\end{document}